\documentclass[10pt,twocolumn,letterpaper]{article}

\usepackage[]{iccv}      
\usepackage{graphicx}
\usepackage{amsmath}
\usepackage{amssymb}
\usepackage{booktabs}
\usepackage[utf8]{inputenc} 
\usepackage[T1]{fontenc}    
\usepackage{url}            
\usepackage{booktabs}       
\usepackage{amsfonts}       
\usepackage{nicefrac}       
\usepackage{microtype}      
\usepackage{xcolor}         
\usepackage{amsmath,amssymb}
\usepackage{graphicx}
\usepackage{mathtools}
\usepackage{algorithm}
\usepackage{algorithmic}
\usepackage{setspace}
\usepackage{amsthm}
\usepackage{booktabs}  
\usepackage{arydshln}
\usepackage{multirow}
\usepackage{url}
\usepackage{graphicx} 
\usepackage{graphics} 
\usepackage{booktabs} 
\usepackage{multirow} 
\usepackage{mathrsfs} 
\usepackage{amsmath} 
\usepackage{booktabs} 
\usepackage{bm}
\definecolor{iccvblue}{rgb}{0.21,0.49,0.74}
\usepackage[pagebackref,breaklinks,colorlinks,allcolors=iccvblue]{hyperref}

\def\paperID{7273} 
\def\confName{ICCV}
\def\confYear{2025}

\title{Text-to-3D Generation by 2D Editing}

\author{Haoran Li\textsuperscript{1}, Yuli Tian\textsuperscript{1}, Yonghui Wang\textsuperscript{1}, Yong Liao\textsuperscript{1} \thanks{Corresponding author}, Lin Wang\textsuperscript{2}, Yuyang Wang\textsuperscript{3}, Peng Yuan Zhou\textsuperscript{4}\\\textsuperscript{1}University of Science and Technology of China, \textsuperscript{2}Nanyang Technological University\\ \textsuperscript{3} The Hong Kong University of Science and Technology (Guangzhou), \textsuperscript{4}Aarhus University \\
{\tt\small \{lhr123, yltian, wyh1998\}@mail.ustc.edu.cn, yliao@ustc.edu.cn} \\
{\tt\small eee-addison.wang@ntu.edu.sg, yuyangwang@hkust-gz.edu.cn, pengyuan.zhou@ece.au.dk  }}

\newcommand{\sysname}{GE3D}
\begin{document}

\twocolumn[{
\renewcommand\twocolumn[1][]{#1}
\maketitle

\begin{center}
    \captionsetup{type=figure}
    \includegraphics[width=\textwidth]{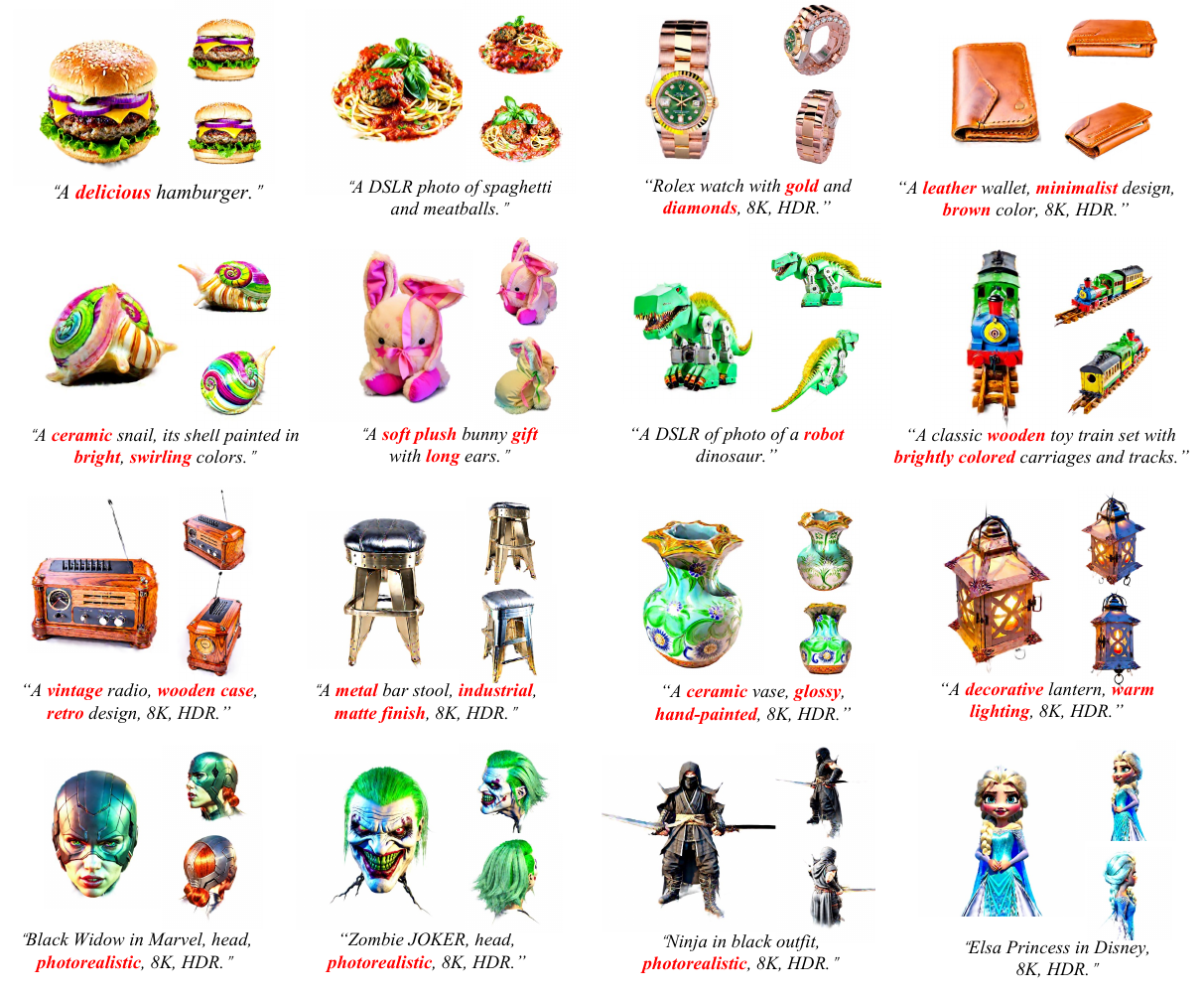}
    \captionof{figure}{The text-to-3D generation results of our framework \textbf{\textit{\sysname}}. \sysname\ replaced the inefficient single-step editing in SDS~\cite{poole2022dreamfusion} with multi-step 2D diffusion editing approach. This change mitigates the issues of over-saturation and over-smoothing, enriching image content and increasing diversity, thereby achieving photorealistic generation quality. Please zoom in to see the details.}
    \label{fig:teaser}
\end{center}
}]

\let\thefootnote\relax\footnotetext{* Corresponding author.}
\begin{abstract}

Distilling 3D representations from pretrained 2D diffusion models is essential for 3D creative applications across gaming, film, and interior design. Current SDS-based methods are hindered by inefficient information distillation from diffusion models, which prevents the creation of photorealistic 3D contents. In this paper, we first reevaluate the SDS approach by analyzing its fundamental nature as a basic image editing process that commonly results in over-saturation, over-smoothing, lack of rich content and diversity due to the poor-quality single-step denoising. In light of this, we then propose a novel method called 3D Generation by Editing (\sysname). Each iteration of \sysname\ utilizes  a 2D editing framework that combines a noising trajectory to preserve the information of the input image, alongside a text-guided denoising trajectory. We optimize the process by aligning the latents across both trajectories. This approach fully exploits pretrained diffusion models to distill multi-granularity information through multiple denoising steps, resulting in photorealistic 3D outputs. Both theoretical and experimental results confirm the effectiveness of our approach, which not only advances 3D  generation technology  but also establishes a novel connection between 3D generation and 2D editing. This could potentially inspire further research in the field. \textbf{\textit{Code and demos are released at }} \url{https://jahnsonblack.github.io/GE3D/}.
\end{abstract}

\section{Introduction}
\label{sec:intro}


\par Distilling 3D representations~\cite{poole2022dreamfusion,lin2023magic3d,wang2024prolificdreamer,metzer2023latent,liang2023luciddreamer,tang2023dreamgaussian,li2024dreamscene} from pretrained 2D text-to-image models~\cite{rombach2022high} has garnered significant attention  in the field of AI-generated content (AIGC). This method enables the generation of 3D representations from scratch based on user-provided text prompts, substantially reducing the workload for 3D modelers. Demonstrating vast potential, this technology promises wide-ranging applications across various industries, such as gaming and augmented/virtual reality (AR/VR).
\par Current research methodologies primarily adhere to Score Distillation Sampling (SDS)~\cite{poole2022dreamfusion}, a pioneering approach in the field of text-to-3D generation. This technique distills knowledge by matching the distribution of images rendered from 3D representations under random poses with that of a pretrained 2D large text-to-image diffusion model~\cite{song2020denoising,rombach2022high}, such as Stable Diffusion~\cite{rombach2022high}, under specific textual contexts. In the SDS process, noise is initially introduced to a rendered image in a single step.  
This noised image is then input into a diffusion model for denoising. The noise predicted by the diffusion model contains information derived from the input text. By reducing the discrepancy between the input and predicted noises, the 3D representation is optimized from the rendered viewpoint. Subsequent advancements in SDS-based methods~\cite{lin2023magic3d,wang2024prolificdreamer,metzer2023latent,liang2023luciddreamer,tang2023dreamgaussian,li2024dreamscene} for 3D generation have introduced a range of  enhancements. These improvements include more advanced 3D representations~\cite{lin2023magic3d,tang2023dreamgaussian}, optimized diffusion timesteps for sampling~\cite{huang2023dreamtime,li2024dreamscene}, and refined techniques for adding noise~\cite{liang2023luciddreamer} in the SDS process.  Despite these developments, challenges in generation quality, such as \textbf{\textit{over-saturation, over-smoothing}}, \textbf{\textit{lack of content richness}} and \textbf{\textit{and lack of diversity}}, persist.
\par
\begin{figure}
\centering
\includegraphics[width=0.45\textwidth]{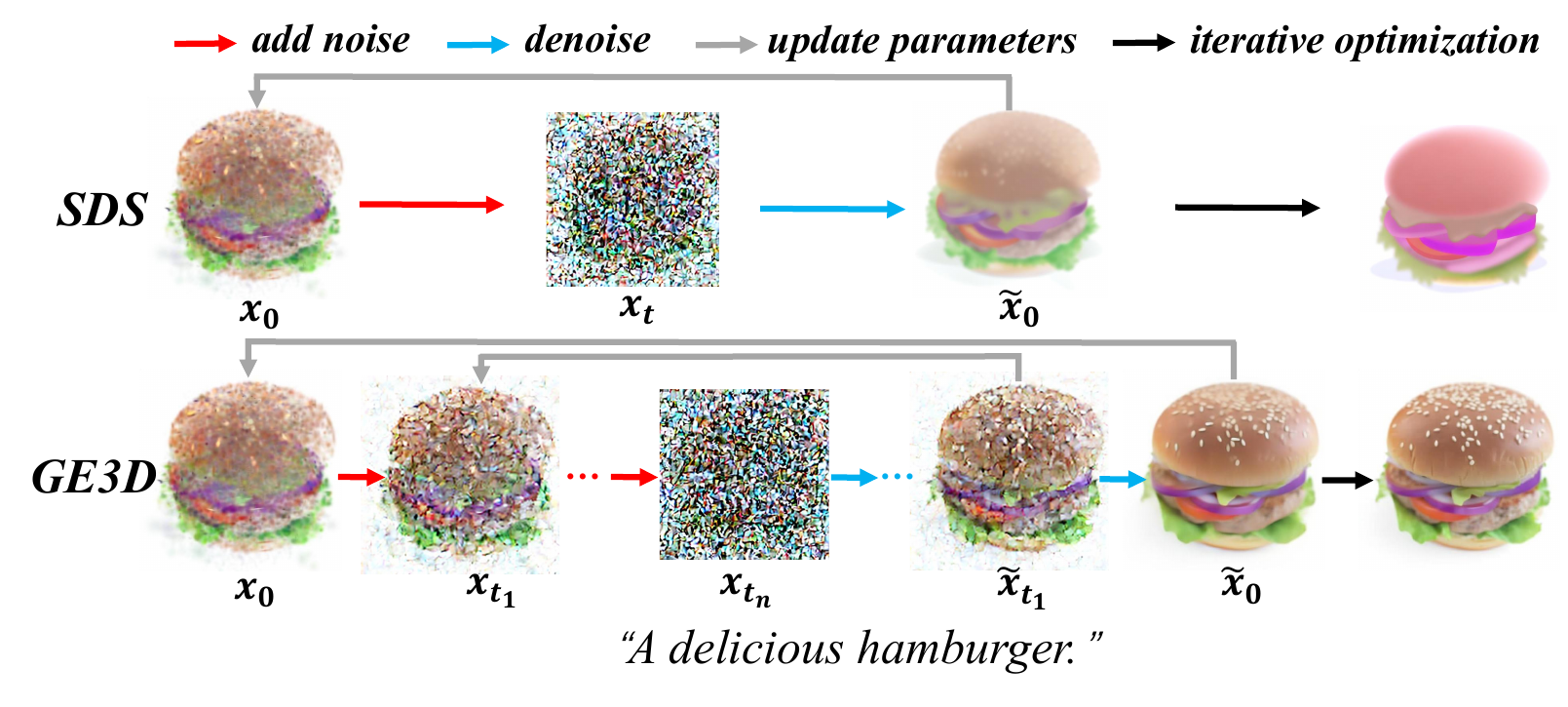}
\centering

\caption{We simulate the 3D generation processes of SDS~\cite{poole2022dreamfusion} and \sysname\ by inputting 2D coarse images through multiple iterations of single-step and multi-step editing. Here, $x_0$
  is the input image, $\tilde{x}_0$
  is the predicted image, and $x_t$ and $\tilde{x}_t$ are the latents along the noising and denoising trajectories, respectively. }
\label{fig:simulate}
\end{figure}

 
We observe that the process used in SDS~\cite{poole2022dreamfusion}, which integrates information from pretrained 2D diffusion models  through noising and denoising, bears similarity to 2D image editing~\cite{hertz2022prompt,mokady2023null,kawar2023imagic}, where the guided input image is methodically altered towards a specified textual direction. The key difference is that SDS employs a \textit{\textbf{single-step}} editing process. However, traditional diffusion models~\cite{song2020denoising, ho2020denoising, rombach2022high} typically require \textit{\textbf{multi-step}} denoising, which involves adding information of varying granularities, to correct directional errors and generate high-quality images. This is because predicting the direction to $x_0$ in a single step often results in a substantial deviation from the true data trajectory. 
As illustrated in Figure~\ref{fig:simulate}, we simulate the SDS single-step image editing process through multiple iterations on a coarse 2D image during the generation process. We observe that 
the content of the images progressively aligns with the textual prompt "\textit{A delicious hamburger}", the quality of the generated images deteriorates: colors become overly saturated and textures become simpler and smoother. \textbf{\textit{This degradation is attributed to significant errors in the predicted generation direction and the introduction of different random noises in each iteration, leading to varying predicted directions.}} These variations make the generation process unstable and tend to drive the outputs towards a \textbf{\textit{simplistic mean direction}} associated with the text. As depicted in the lower section of Figure~\ref{fig:simulate}, high-quality image generation can be achieved by fully exploiting the multi-step denoising capabilities of diffusion models in each optimization iteration, thereby facilitating a comprehensive image editing process.

Based on the analysis, we propose \textbf{\textit{\sysname}}, \textbf{\textit{a novel 3D generation framework designed to produce photorealistic 3D content, which no longer relies on an SDS-based architecture.}} This framework aligns the latents of the input images during the 2D diffusion editing process. In each iteration, we establish a noising trajectory to project  the input image into  high-dimensional diffusion space, accompanied by a denoising editing trajectory that incrementally incorporates information of varying granularities from the diffusion model. Larger timesteps impart coarse-grained information, while smaller timesteps contribute fine-grained details. By minimizing the differences in latents at corresponding timesteps along these trajectories, \sysname\ effectively distills multi-granularity information into the 3D representation. 
 Additionally, we introduce a dynamic balancing coefficient (DBC) related to the number of iterations and denoising steps, optimizing the provision of coarse-grained information in initial phases and fine-grained information in later stages, thereby enhancing stability and efficiency in generation. The efficacy of \sysname\ in achieving photorealistic 3D generation is demonstrated  in Figure~\ref{fig:teaser}.
Our main contributions can be summarized as follows:

\begin{itemize}
\item We conduct a comprehensive analysis of the relationship between 2D editing and 3D generation, demonstrating both qualitatively and quantitatively that the core of the SDS method lies in performing \textbf{\textit{single-step}} editing during each optimization, which introduces significant errors in the predicted direction, ultimately causing the generation to collapse into a simplistic mean result.

\item We introduce a novel text-to-3D generation framework, \sysname. By aligning the latents along the noising and denoising trajectories in each optimization step and integrating the dynamic balancing coefficient, we achieved efficient, high-quality and diverse 3D generation.

\item  Both qualitative and quantitative results showcase the significant superiority of \sysname\ in generating photorealistic 3D representations that closely match text descriptions. 
Our work could be potential to inspire further research that integrates advanced 2D editing techniques into the realm of 3D generation.

\end{itemize}
\section{Related Works}
\label{sec:related}
\subsection{Text-to-3D Generation}
Current text-to-3D  generation methods can be broadly  categorized  into two primary approaches: direct generation~\cite{nichol2022point,jun2023shap,shi2023mvdream, hong2023lrm,yang2024tencent} and optimization-based generation~\cite{poole2022dreamfusion,lin2023magic3d,chen2023fantasia3d,metzer2023latent,liang2023luciddreamer,tang2023dreamgaussian,li2024dreamscene, lukoianov2024score}. Direct generation methods, such as Point-E~\cite{nichol2022point}, utilize  a diffusion model to generate an image corresponding to the input text, subsequently training a transformation model to convert this image into a point cloud with a predetermined number of points. Alternatively, models like LRM~\cite{hong2023lrm}, employ encoders to map text or images into 3D high-dimensional spaces, which are then decoded into specific 3D representations, such as meshes or 3D Gaussians~\cite{kerbl20233d}. While these methods~\cite{hong2023lrm,zhao2025hunyuan3d,nichol2022point} can rapidly generating 3D representations, they involve high training costs and typically yield simpler 3D structures, with the diversity of the generated objects depending on the dataset, which may not be suitable for complex 3D applications.

\par Optimization-based methods are capable of producing higher quality and more diverse 3D representations but generally require longer optimization times. Score Distillation Sampling (SDS)~\cite{poole2022dreamfusion}, a seminal work in this field, optimizes the process to ensure that images rendered from any camera pose conform to the distribution of a large pretrained text-to-image model. Building on SDS, subsequent research such as CSD~\cite{yu2023text} decomposes the SDS loss into classification and generative components, noting improvements in quality with generative loss alone. ProlificDreamer~\cite{wang2024prolificdreamer} introduced Variational Score Distillation (VSD), which treats 3D parameters as random variables and refines the distillation through particle sampling for model diversification. Consistant3D~\cite{wu2024consistent3d} utilizes fixed Gaussian noise to maintain consistency in the ordinary differential equation (ODE) sampling process. LucidDreamer~\cite{liang2023luciddreamer} employs the DDIM~\cite{song2020denoising} inversion method to enhance the consistency and quality of the optimization process. Further, DreamTime~\cite{huang2023dreamtime} demonstrate that different diffusion timesteps in a pretrained model offer variable information, with specific time sampling strategies significantly improving both quality and speed of generation. However, despite these advancements, these methods still struggle to produce photorealistic 3D content.
\par These two types of  approaches remain unviable for practical applications due to limitations in generation quality.

\subsection{Text-based Image Editing}
\label{sec:related-editing}
Current text-guided image editing methods generally involve two main stages. The first is image inversion, which involves mapping the image back to the latent space of a generative model (such as GANs~\cite{karras2020analyzing,goodfellow2020generative} or diffusion models~\cite{ho2020denoising,song2020denoising,rombach2022high}) through a reconstruction loss.  The second 
is editing that the latents is guided by a new text in latent space and achieves final modifications in image space while preserving most of the original content. For GAN-based methods~\cite{tov2021designing,alaluf2022hyperstyle,wang2022high}, an encoder~\cite{tov2021designing,wei2022e2style} is typically trained to map images to latents, and the latents are precisely obtained through optimization techniques~\cite{li20253d}. For methods based on diffusion models, such as Imagic~\cite{kawar2023imagic}, precise reconstruction is achieved by optimizing the alignment of the target text embedding with the image embedding, though this usually requires retraining the diffusion model at a high cost.

In advanced techniques, tuning-free methods like DDIM Inversion~\cite{song2020denoising} can approximate the inversion trajectory through deterministic sampling, but this method often results in reconstruction images that deviate from the input image, thereby affecting subsequent editing. Null-text Inversion(NTI)~\cite{mokady2023null} optimizes a differentiable empty text embedding to align the noising and denoising processes for more precise reconstruction. Meanwhile, Negative Prompt Inversion(NPI)~\cite{miyake2023negative} eliminates the optimization process used in NTI and directly uses the prompt of input image as an approximate substitute, which is faster but gets some reconstruction errors. Prompt Tuning Inversion(PTI)~\cite{dong2023prompt} achieves precise reconstruction without explicitly inputting the image prompt by optimizing the prompt  during the denoising process. Once the deterministic diffusion trajectories are obtained, image editing based on new text can be completed through interpolation of text embeddings during the generation process or using the Prompt2Prompt~\cite{hertz2022prompt} technique.

\section{Methods}
In this section, we first introduce the overall process of diffusion-based 2D editing and conduct both qualitative and quantitative analyses of its relationship with SDS-based 3D generation methods. Then, we explore the limitations of existing SDS-based approaches, identifying their deficiencies as stemming from the single-step denoising process. Finally, we propose \sysname, which enables stable and high-quality 3D generation. Due to space limitations, we have placed the relevant preliminary knowledge, additional comparisons between 2D editing methods and 3D methods, and further details of \sysname\ in Section~\ref{sec:supp_preliminay}, Section~\ref{sec:supp_2dvs3d} and Section~\ref{sec:supp_adaptable} of the Supplementary Materials, respectively. 
\subsection{From 2D editing to SDS-based 3D generation}
Text-based 2D editing modifies an input image $x$ based on a target text $y$
 , essentially treating it as an image-to-image task. Typically, the process involves mapping $x$ into a latent space, editing its features according to $y$
 , and decoding the modified features back into an image. This preserves non-target content while only altering the text-related information. Diffusion models~\cite{ho2020denoising,song2020denoising,rombach2022high} have become dominant due to their Markovian noising and denoising mechanism, which simplifies latent mapping and feature regeneration. The noising process in diffusion models is as follows:
\begin{equation}
    x_t = \sqrt{\bar{\alpha_t}}x + \sqrt{1-\bar{\alpha_t}}\epsilon . 
\label{eq:ddpm}
\end{equation}
where $x_t$ represents the input $x$  adding $t$-step noise $\epsilon$ and $\overline{\alpha}_t$ is a coefficient corresponding to $t$.
\par As discussed in Section~\ref{sec:related-editing}, the editing process consists of two key stages: image inversion and text embedding modification. The purpose of image inversion is to get a specific generation trajectory that corresponds to the given input image $x$. We define a diffusion trajectory for noising as $\{{x_{t_0},x_{t_1},...,x_{t_{n-1}},x_{t_n}}\}$ ( $x=x_{t_0},t_n>t_{n-1}>...>t_1>t_0$, ) and a denoising trajectory $\{\tilde{x}_{t_{n-1}},...,\tilde{x}_{t_1},\tilde{x}_{t_0}\}$. We use the following DDIM sampling~\cite{song2020denoising} process to perform noising  and denoising for adjacent trajectory nodes $x_p$ and $x_q$ (for simplicity, we use $p$ and $q$ to represent different time steps and no matter $p>q$ or $p<q$):
\begin{equation}
\resizebox{0.9\linewidth}{!}{
    $x_p = \sqrt{\overline{\alpha}_p} \frac{x_q - \sqrt{1-\overline{\alpha}_q}\epsilon_{\phi}(x_q,q)}{ \sqrt{\overline{\alpha}_q}} + \sqrt{1-\overline{\alpha}_p}\epsilon_{\phi}(x_q,q).$}
\label{eq:ddim}
\end{equation}
where $\phi$ is a pre-trained noise prediction network. We need to progressively align the corresponding latents $x_{t_i}$ and $\tilde{x}_{t_i}$($i=0,...,n-1$) to achieve inversion. Methods such as NTI~\cite{mokady2023null} and PTI~\cite{dong2023prompt} use differentiable embeddings for the unconditional empty text $ \emptyset_{t}$ or the conditional text $y_{t}$ during the denoising process, enabling optimization for alignment. We represent this process as:   
\begin{equation}
    \alpha_{t_i} = \arg \min\limits_{ \alpha_{t_i}} ||x_{t_{i-1}} - \tilde{x}_{t_{i-1}}(\tilde{x}_{t_i},t_i, \alpha_{t_i})||_2^2,
\label{eq:reconstruction}
\end{equation}
where $i=n,..., 1$ and $\alpha_{t_i} = \emptyset_{t_i}$ or $y_{t_i}$. And after this alignment process, the new text embedding $y_{edit}$ can be combined with the optimized embedding sequence $\alpha_{t_i}$ ($i=1,...,n$)  to perform editing.

\par  \textbf{Our idea:} Thus, we can view the two stages of 2D editing as a process where an image initially preserves its inherent features and then gradually shifts toward the target text. We hypothesize that 3D content generation follows a similar stable progression. For example, in SDS-based methods~\cite{poole2022dreamfusion,lin2023magic3d,chen2023fantasia3d,metzer2023latent,liang2023luciddreamer,tang2023dreamgaussian,li2024dreamscene,lukoianov2024score}, when the rendering camera viewpoint remains fixed, the image transitions from an initial blurred state to a clear depiction of the object described by the target text. Qualitatively, this evolution resembles the style transformation in 2D editing—albeit with much more extensive content variation in 3D. If we consider a single iteration in the 3D generation process, it essentially constitutes a simplified version of 2D editing. The quantitative proof is as follows:
\par For SDS, similarly to many method analyses~\cite{liang2023luciddreamer,huang2023dreamtime}, we directly add noise using Eq.~\ref{eq:ddpm} to $x_{t_0}$ across $t_n$-timestep and then use Eq.~\ref{eq:ddim} for denoising in just one step to get $\tilde{x}_{t_0}$, we can derive the following formula:
\begin{equation}
\left\{
\begin{aligned}
x_{t_n} &= \sqrt{\bar{\alpha}_{t_n}}x_{t_0} + \sqrt{1-\bar{\alpha}_{t_n}}\epsilon 
  \\
\tilde{x}_{t_0} &= \frac{x_{t_n} - \sqrt{1-\overline{\alpha}_{t_n}}\tilde{\epsilon}_{\phi}(x_{t_n},t_n,y,\emptyset)}{\sqrt{\overline{\alpha}_{t_n}}}.
\end{aligned}
\right.
\end{equation}
where $\tilde{\epsilon}_{\phi}(x_t,t,y,\emptyset) =  \epsilon_{\phi}(x_t,t,\phi) + \lambda(\epsilon_{\phi}(x_t,t,y) - \epsilon_{\phi}(x_t,t,\emptyset))$, $\lambda$ is the classifier-free guidance (CFG) coefficient. By  eliminating $x_{t_n}$, we can obtain:
\begin{equation}
    \sqrt{\frac{\overline{\alpha}_{t_n}}{1-\overline{\alpha}_{t_n}}} (x_{t_0} - \tilde{x}_{t_0}) = \tilde{\epsilon}_{\phi}(x_{t_n},t_n,y,\emptyset) - \epsilon,
\label{eq:re_SDS}
\end{equation}
where $\tilde{\epsilon}_{\phi}(x_{t_n},t_n,y,\emptyset) - \epsilon$ is the core of SDS.

\par LucidDreamer~\cite{liang2023luciddreamer} is currently the state-of-the-art (SOTA) method. Its ISM replaces $\tilde{\epsilon}_{\phi}(x_t, t, y, \emptyset) - \epsilon$ in SDS with $\tilde{\epsilon}_{\phi}(x_t, t, y, \emptyset) - \epsilon_{\phi}(x_s,s,\emptyset)$ where $x_t$ is obtained by adding noise to $x_s$ using DDIM Inversion~\cite{song2020denoising}. We set $t_{n-1} = s$, $ t_n = t$ and add noise through $x_{t_{n-1}}$ to get $x_{t_n}$ , and then denoise from $x_{t_n}$ to retrieve $\tilde{x}_{t_{n-1}}$ by Eq.~\ref{eq:ddim}. The equations in this process can be expressed as follows:
\begin{equation}
\left\{
\begin{aligned}
     \frac{x_{t_n}}{\sqrt{\overline{\alpha}_{t_n}}}  = \frac{x_{t_{n-1}}}{\sqrt{\overline{\alpha}_{t_{n-1}}}} &+\delta_{t_n}\epsilon_\phi(x_{t_{n-1}},t_{n-1},\emptyset) \\
       \frac{\tilde{x}_{t_{n-1}}}{\sqrt{\overline{\alpha}_{t_{n-1}}}} =\frac{x_{t_n}}{\sqrt{\overline{\alpha}_{t_n}}}&- \delta_{t_n}\tilde{\epsilon}_\phi(\tilde{x}_{t_n},t_n,y,\emptyset),
\end{aligned}
\right.
\end{equation}
where $\delta_{t_n} = \sqrt{\frac{1-\overline{\alpha}_{t_n}}{\overline{\alpha}_{t_n}}}-\sqrt{\frac{1-\overline{\alpha}_{t_{n-1}}}{\overline{\alpha}_{t_{n-1}}}}$. By adding the two equations and eliminating $x_{t_n}$, we can obtain:
\begin{equation}
\resizebox{0.88\linewidth}{!}{$
     \frac{x_{t_{n-1}}-\tilde{x}_{t_{n-1}}}{\sqrt{\overline{\alpha}_{t_{n-1}}}\cdot\delta_{t_n}}=  \tilde{\epsilon}_\phi(\tilde{x}_{t_n},t_n,y,\emptyset) - \epsilon_\phi(x_{t_{n-1}},t_{n-1},\emptyset).$}
\label{eq:re_ISM}
\end{equation}
Thus, based on Eq.~\ref{eq:reconstruction}, Eq.~\ref{eq:re_SDS} and Eq.~\ref{eq:re_ISM}, we can observe that SDS and ISM  essentially align the latents $x_{t_i}$($i=0$ or $n-1$) at different timesteps along the diffusion trajectory of inversion process in 2D editing task. 


\subsection{Why SDS-based Methods Fail: An Analysis}
\label{sec:why}
Although SDS-based methods have been proven to be related to 2D editing, they fail to achieve the stable and high-quality generation process we hypothesized, primarily due to their \textbf{\textit{single-step}} denoising approach. Traditional diffusion models~\cite{ho2020denoising,song2020denoising,rombach2022high} rely on \textbf{\textit{multi-step}} denoising, which integrates information at different granularities for high-quality generation. During the early denoising stages, the model captures coarse-grained details (e.g., position and general shape of objects), while later stages refine fine-grained details (e.g., object edges). Consequently, in 2D editing tasks, aligning information across multiple denoising steps is crucial for achieving accurate image inversion and ensuring high-quality image editing.

\par However, in the case of SDS, the single-step denoising process predicts the target direction of $x_{t_0}$ inaccurately, resulting in significant errors. Additionally, since its noising process does not incorporate DDIM Inversion to retain part of the current 3D content, the generation process becomes unstable, causing the results to collapse toward the mean and lose diversity, as seen in Figure~\ref{fig:simulate}. For ISM~\cite{liang2023luciddreamer}, although it utilizes DDIM Inversion to preserve some 3D content, its single-step denoising process 
$\frac{\tilde{x}_{t_{n-1}}}{\sqrt{\overline{\alpha}_{t_{n-1}}}} =\frac{x_{t_n}}{\sqrt{\overline{\alpha}_{t_n}}}- \delta_{t_n}\tilde{\epsilon}_\phi(\tilde{x}_{t_n},t_n,y,\emptyset)$ only aligns  $x_{t_{n-1}}$
  with $\tilde{x}_{t_{n-1}}$
 , retaining only coarse-grained information. As observed in the Figure~\ref{fig:compare} and Figure~\ref{fig:supp_lucid} in Supplementary Material, this leads to a significant lack of fine-grained details, preventing the generation of photorealistic quality. Therefore, we need to introduce a multi-step denoising process into 3D optimization to achieve stable and high-quality generation.

\begin{figure}
\centering
\includegraphics[width=0.49\textwidth]{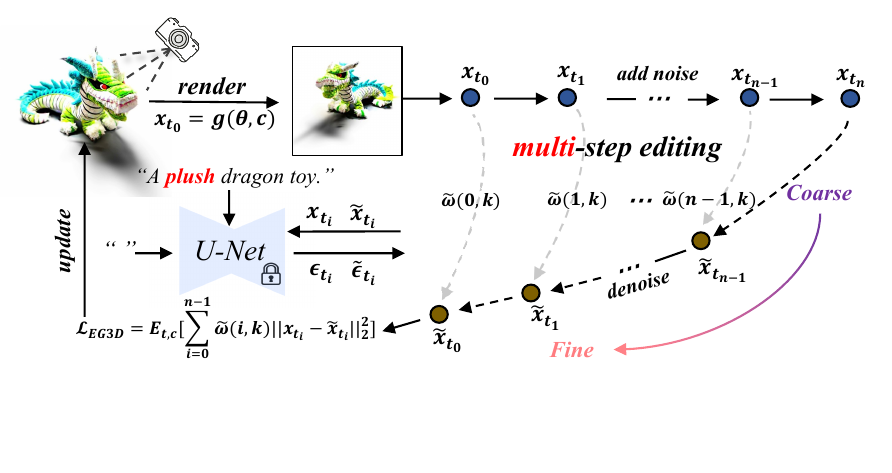}
\centering
\vspace{-40pt}
\caption{The overview of \sysname. We integrated the 2D editing process into 3D generation. Unlike the single-step editing in SDS in DreamFusion~\cite{poole2022dreamfusion} and ISM in LucidDreamer~\cite{liang2023luciddreamer}, we used multi-step editing with latents alignment to combine different granularities of information from the pre-trained 2D diffusion model into the 3D representation, achieving high-quality generation.}
\label{fig:overview}
\end{figure}

\subsection{The Proposed \sysname ~Framework}
\label{sec:GE3D}
Based on above analysis, we find that directly integrating the 2D editing framework into 3D generation presents two main challenges: 1) 2D editing involves image inversion and text embedding modification, with image inversion requiring multiple optimization steps. Integrating this into 3D generation would be too time-consuming, and 2) 2D editing follows a stable refinement from coarse to fine details, but achieving this in 3D optimization is highly difficult.



\par First, we discard the step-by-step optimization in 2D editing and directly align image and text-edited latents using L2 loss. As seen in Figure~\ref{fig:overview}, similar to 2D editing tasks, we define an $n$-step noising trajectory and a corresponding reverse denoising trajectory. During the noising process, we employ an ODE-guided DDIM process with empty text embeddings to preserve the features of the input image. In the denoising process, we use the guidance information calculated by CFG and obtain a sequence of semantic information with varying granularities through the stepwise denoising process. We then use $L_2$ loss to align the corresponding latents along the two trajectories:
\begin{equation}
\setlength{\abovedisplayskip}{1pt}
\setlength{\belowdisplayskip}{1pt}
    L_{i} = ||x_{t_i}-\tilde{x}_{t_i}||_2^2, i=0,...,n-1. 
\end{equation}
This  forces the 3D representation to learn rich semantic information at different granularities from the current viewpoint, resulting in the generation of high-quality 3D content.

\begin{figure*}[t!]
\centering
\includegraphics[width=0.95\textwidth]{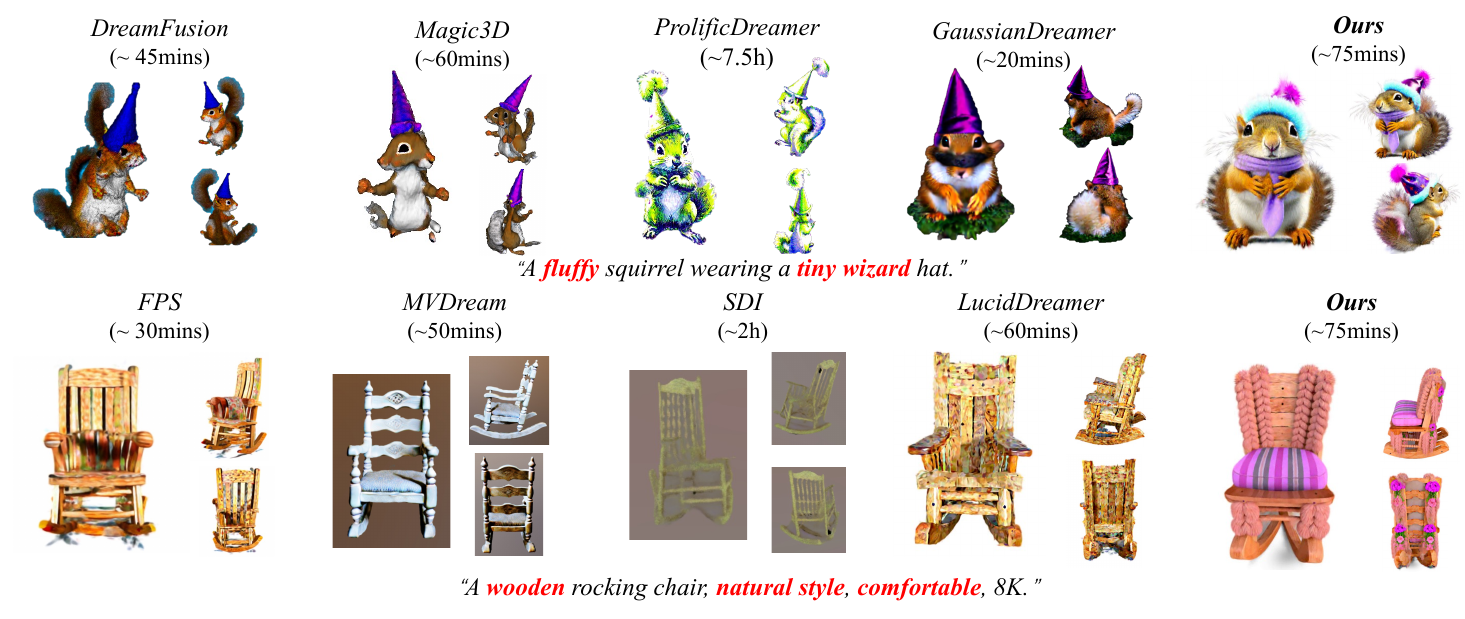}
\centering
\caption{Qualitative comparison with baselines in text-to-3D generation.}
\label{fig:compare}
\end{figure*}
\par This alignment of information at different granularities can \textbf{\textit{accelerate the convergence speed}} of 3D representation generation. Although multi-step denoising is more time-consuming than single-step denoising, it requires fewer iterations. As a result, \sysname\ achieves significantly superior generation quality compared to current SOTA methods without a noticeable increase in generation time.

\par Then, to achieve a stable coarse-to-fine optimization process, we propose a \textbf{\textit{dynamic balancing coefficient (DBC)} $\omega$} to balance information at different granularities across varying iteration steps. We use $\omega(i,k)$ to represent the coefficient applied to the loss $L_i$ in the $k$-th iteration. We aim for a higher $\omega(i,k)$ contribution from coarse-grained losses $L_i$ (large $i$) when iteration $k$ is small, as the shape of 3D representation has not yet been optimized, and overly fine-grained information (small $i$)  could interfere its rich  semantic information. During the early stages of optimization, coarse-grained information is primarily used. As the optimization progresses, the contribution of $L_i$ (large $i$) becomes less beneficial for 3D generation, and more fine-grained information 
 (small $i$) is needed to achieve high-quality results.
\par We use multiple Gaussian functions to model the characteristic where the coefficient first increases and then decreases. For each 
$\omega(i,k) = \exp{(-\frac{(k-\Delta_i)^2}{2\sigma^2})}$ where $\Delta_i$ represents the iteration number at which the peak occurs of $i$-th Gaussian function. So, for the endpoints $\Delta_{n-1} = 0$ and $\Delta_{0} = K$ ($K$ denotes the total number of iterations). We set $\Delta_i$ at equal intervals as follows:
\begin{equation}
\setlength{\abovedisplayskip}{1pt}
\setlength{\belowdisplayskip}{1pt}
    \Delta_i = (n-i-1)\times\frac{K}{n-1},i=0,...,n-1.
\end{equation}
We  set a large value ($1000$) for $\sigma$ to ensure that information at different granularities participates in the optimization process within each iteration. Then we normalize all the coefficients $\omega_{i,k}$ at the same iteration $k$ as follows:
\begin{equation}
\setlength{\abovedisplayskip}{1pt}
\setlength{\belowdisplayskip}{1pt}
    \tilde{\omega}(i,k) = \frac{\omega(i,k)}{\sum_{i=0}^{n-1}\omega(i,k)}.
\label{eq:normalization}
\end{equation}
Thus, the  loss for \sysname\ can be expressed as:
\begin{equation}
\setlength{\abovedisplayskip}{1pt}
\setlength{\belowdisplayskip}{1pt}
    L_{\sysname} = \mathbb{E}_{t,c}\left[ \sum_{i=0}^{n-1}  \tilde{\omega}(i,k)||x_{t_i}-\tilde{x}_{t_i}||_2^2 \right].
\end{equation}
A more detailed analysis of DBC is provided in Section~\ref{sec:supp_DBC} of the Supplementary Material.
\par Our complete algorithmic process is shown in Algorithm~\ref{alg:GE3D} and Figure~\ref{fig:overview}.
\begin{algorithm}[] 
\caption{GE3D}
\vspace{-2pt}
\label{alg:GE3D}
\begin{algorithmic}[1]
\STATE \textbf{Input:} diffusion timestep trajectory$\{t_0,t_1,...,t_n\}$, renderer $g$, camera pose $c$, 3D representation parameters 
$\theta$, iteration number $K$, target text embedding $y$, and null text embedding $\emptyset$.
\FOR{$k = [0, ..., K-1]$}
\STATE $x_{t_0} = g(\theta, c)$
\FOR{$i = [0, ..., n-1]$} 
\STATE  $\Delta_i = (n-i-1)\times\frac{K}{n-1}$
\STATE  $\omega(i,k) = \exp{(-\frac{(k-\Delta_i)^2}{2\sigma^2})}$
\STATE  $\tilde{\omega}(i,k) = \frac{\omega(i,k)}{\sum_{i=0}^{n-1}\omega(i,k)}$
\STATE $\epsilon_\phi(x_{t_{i}},t_{i},\emptyset) =$ U-Net$(x_{t_i},t_i,\emptyset)$
\STATE \resizebox{0.55\linewidth}{!}{$\delta_{t_{i+1}} = \sqrt{\frac{1-\overline{\alpha}_{t_{i+1}}}{\overline{\alpha}_{t_{i+1}}}}-\sqrt{\frac{1-\overline{\alpha}_{t_{i}}}{\overline{\alpha}_{t_{i}}}}$}
\STATE $x_{t_{i+1}} = \sqrt{\overline{\alpha}_{t_{i+1}}} 
 (\frac{x_{t_{i}}}{\sqrt{\overline{\alpha}_{t_{i}}}} +\delta_{t_{i+1}}\epsilon_\phi(x_{t_{i}},t_{i},\emptyset))$
\ENDFOR
\STATE $\tilde{x}_{t_{n}}= x_{t_n}$ 
\FOR{$i = [n-1, ..., 0]$}

\STATE $\tilde{\epsilon}_\phi(\tilde{x}_{t_{i+1}},t_{i+1},y,\emptyset) =$ U-Net$(\tilde{x}_{t_{i+1}},t_{i+1},y,\emptyset)$
\STATE $\tilde{x}_{t_{i}}=\sqrt{\overline{\alpha}_{t_{i}}}(\frac{\tilde{x}_{t_{i+1}}}{\sqrt{\overline{\alpha}_{t_{i+1}}}}- \delta_{t_{i+1}}\tilde{\epsilon}_\phi(\tilde{x}_{t_{i+1}},t_{i+1},y,\emptyset))$
\ENDFOR
\STATE $\nabla_{\theta}L_{\sysname} =  \sum_{i=0}^{n-1}  \tilde{\omega}(i,k)(x_{t_i}-\tilde{x}_{t_i})$
\STATE Update $\theta$
\ENDFOR
\end{algorithmic}
\vspace{-2pt}
\end{algorithm}

 \noindent \textbf{Diversity}. Through this multi-step structured optimization, information gradually transfers from coarse-to-fine granularity in the 3D representation, enabling a generation process that forms geometry and refines textures. This structured pipeline prevents collapse into an average result, as seen in SDS-based methods, and instead enhances texture diversity. Even with the same random seed, our approach preserves geometric consistency while introducing texture variations.

 \noindent \textbf{Easily adaptable}. Our method uncovers the essence of SDS by using a single-step denoising process for coarse editing. In this sense, SDS can be viewed as a subset of our approach, meaning that many improvements and techniques developed for SDS can be seamlessly applied to ours. For instance, we can directly integrate the denoising formula from Perp-Neg~\cite{armandpour2023re} to mitigate the Janus (multi-head) problem. Similarly, we can replace the single-step denoising in Reconstructive Generation from Formation Pattern Sampling~\cite{li2024dreamscene} with multi-step editing to accelerate surface texture generation. 

\section{Experiments}
\par \noindent \textbf{Implementation Details.} Our method is applicable to multiple 3D representations~\cite{mildenhall2021nerf,shen2021deep, kerbl20233d}. In this paper, we use 3D Gaussians~\cite{kerbl20233d}, an explicit representation with strong application potential. We utilized Point-E~\cite{nichol2022point} to generate initial sparse Gaussian point clouds of objects. For the 2D text-to-image diffusion models, we employed Stable Diffusion 2.1~\cite{rombach2022high}. In our diffusion trajectory,  we defined time nodes ranging from $5$ to $10$, with step intervals between nodes set at $40$ to $60$, $60$ to $80$, and $80$ to $100$. We set the  classifier-free guidance (CFG) coefficient $\lambda$ to $7.5$. To ensure a fair comparison, we conducted tests for \sysname\ and all baseline methods using the same NVIDIA A6000 GPU.
\par \noindent \textbf{Baselines.} For the comparative analysis of text-to-3D generation, we utilize the current open-sourced state-of-the-art(SOTA) methods as our baselines: DreamFusion~\cite{poole2022dreamfusion}, Magic3D~\cite{lin2023magic3d}, ProlificDreamer~\cite{wang2024prolificdreamer}, GaussianDreamer~\cite{yi2023gaussiandreamer}, FPS~\cite{li2024dreamscene}, MVDream~\cite{shi2023mvdream}, SDI~\cite{lukoianov2024score} and LucidDreamer~\cite{liang2023luciddreamer}.
(DreamFusion, ProlificDreamer, Magic3D and SDI have been reimplemented by Three-studio~\cite{guo2023threestudio}).
\par \noindent  \textbf{Evaluation Metrics.}  We use CLIP similarity~\cite{radford2021learning} to measure the alignment between the rendered image of the generated 3D representation and and their corresponding text descriptions. Additionally, for assessing the quality of generated 3D representations, we calculate the  Fréchet Inception Distance (FID)~\cite{heusel2017gans} between the rendered images and real-world images, as well as using the Blind/Referenceless Image Spatial Quality Evaluator (BRISQUE)~\cite{mittal2012no}. For all quantitative comparison experiments, we generated 100 3D representations for each text prompt across all baselines, randomly sampling 20 images from each set and computing the aforementioned metrics.
\begin{table}[]
\caption{Quantitative results of \sysname\ compared with baselines. $\uparrow$ means the larger the better while $\downarrow$ means the smaller the better.}
\label{tab:compare}
\resizebox{0.48\textwidth}{17mm}{
\begin{tabular}{c|c|c|c|c}
\hline
& \multicolumn{2}{c|}{ CLIP Similarity}& \multirow{2}{*}{FID $\downarrow$  }& \multirow{2}{*}{BRISQUE $\downarrow$ } \\ \cline{2-3}
& ViT-L/14 $\uparrow$    & ViT-BigG/14 $\uparrow$     &       &   \\ \hline
Dreamfusion~\cite{poole2022dreamfusion}     &       0.278       &       0.370          & 265.57 &  86.58     \\
Magic3D~\cite{lin2023magic3d}         &        0.262      &     0.351          &    285.20  &  79.74   \\
ProlificDreamer~\cite{wang2024prolificdreamer}   &       0.287       &      0.399          &   254.53      &  74.16  \\
GaussianDreamer~\cite{yi2023gaussiandreamer} &     0.286         &  0.396               &       244.48  &  86.91 \\
FPS~\cite{li2024dreamscene}   &0.286&0.398&221.75&70.64  \\
MVDream~\cite{shi2023mvdream} &0.289&0.401&203.62&62.73 \\
SDI~\cite{lukoianov2024score}   &0.289&0.400&209.34&63.86  \\
LucidDreamer~\cite{liang2023luciddreamer}    &      0.290        &   0.402              &   199.19    &  60.22   \\ \hline
\textbf{Ours}            &         \textbf{0.298}     &       \textbf{0.423}          & \textbf{148.33}& \textbf{46.99}     \\ \hline 
\end{tabular}
}
\end{table}

 \begin{figure}
\centering
\includegraphics[width=0.49\textwidth]{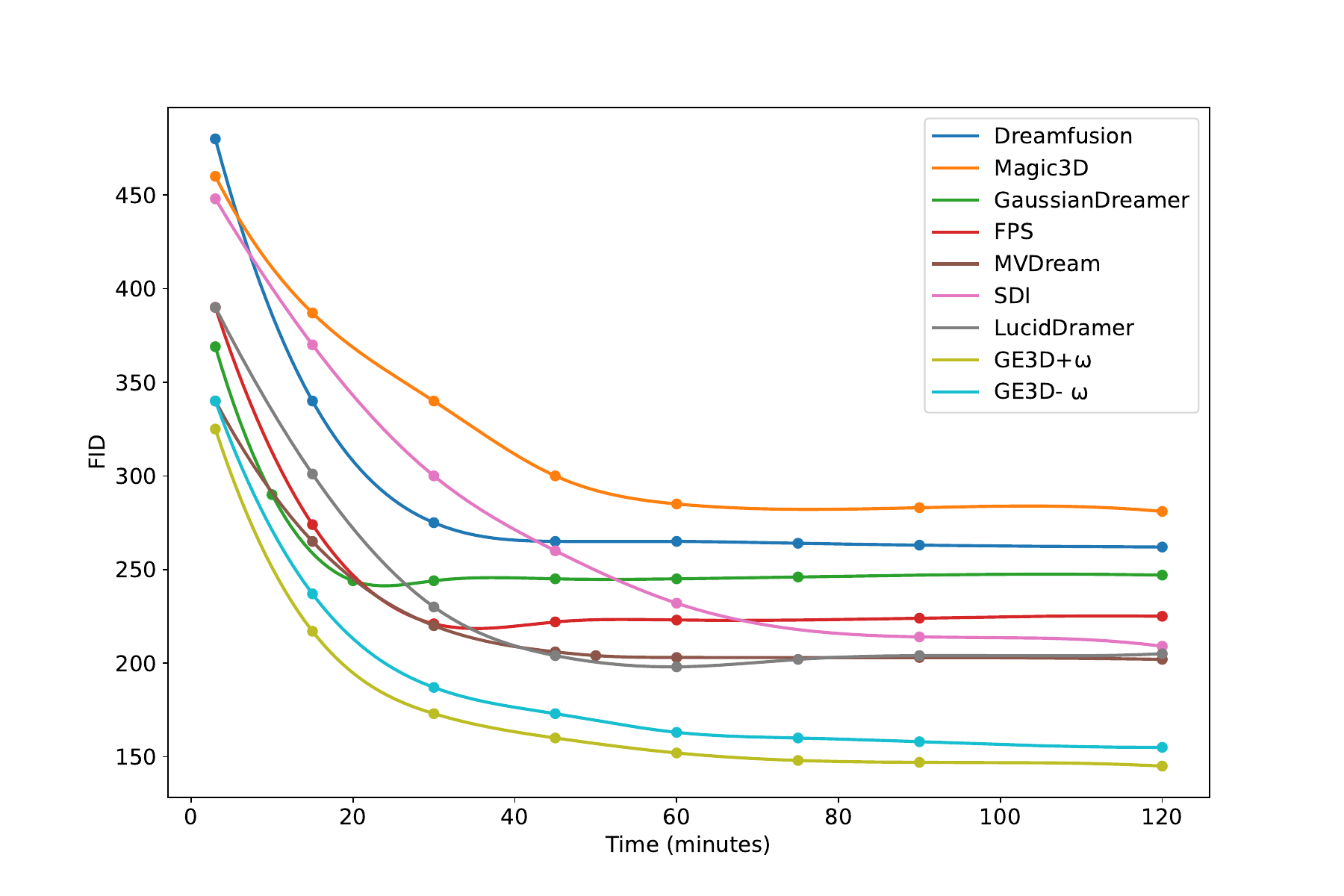}
\centering
\caption{The convergence speed of different text-to-3d methods. Due to the excessive time consumption, we did not plot ProlificDreamer's results.}
\label{fig:convergence}
\end{figure}

\subsection{Qualitative Results}
%

Figure~\ref{fig:teaser} qualitatively showcases the results of \sysname\ across the generation of various types of 3D objects.  By employing a multi-step 2D editing process in each iteration, our framework fully leverages the capabilities of large pretrained 2D diffusion models~\cite{rombach2022high}. 
As shown, \sysname\ not only generates photorealistic results for categories such as animals, luxury goods, and dolls, but it also excels in more challenging tasks, such as generating 3D avatars and human faces, with results that are nearly indistinguishable from real-world images. We provide additional image and video results in Section~\ref{sec:supp_more} of the Supplementary Material and the ./video\_results folder.
\par Figure~\ref{fig:compare} presents a qualitative comparison between our method and other baseline approaches. As demonstrated, our generated 3D representations significantly outperform other methods~\cite{poole2022dreamfusion,lin2023magic3d, wang2024prolificdreamer, yi2023gaussiandreamer, liang2023luciddreamer} in terms of quality. For the prompt "\textit{A fluffy squirrel wearing a tiny wizard hat}", our method produces fur that appears more lifelike and hat details that are clearer, resulting in a more realistic overall appearance.
For the prompt "\textit{A wooden rocking chair, natural style, comfortable, 8K}", our results and those from LucidDreamer~\cite{liang2023luciddreamer}  are closer to the text prompt than those from other methods. However, LucidDreamer's output displays rough textures and many surface artifacts. This issue stems from LucidDreamer’s ISM~\cite{liang2023luciddreamer} method, which relies on single-step editing within a diffusion model, as discussed in Section~\ref{sec:why}.  Consequently, ISM-generated 3D representations lack fine details.  In contrast, our 3D outputs showcase enhanced details, eliminate artifacts, and improve lighting effects, achieving superior photorealistic quality. We provide a frame-by-frame comparison and video comparison results with LucidDreamer in Section~\ref{sec:supp_ISM} and in the ./compare\_ISM folder of the Supplementary Material.

\subsection{Quantitative Results}
In Table \ref{tab:compare}, the CLIP similarity results indicate that our generated 3D representations align more closely with the input text prompts, a benefit derived from the multi-step editing process in each optimization that incorporates richer textual information. Additionally, the quantitative assessments from FID and BRISQUE scores further confirm that our method produces 3D contents of superior generative quality.

\begin{figure}
\centering
\includegraphics[width=0.48\textwidth]{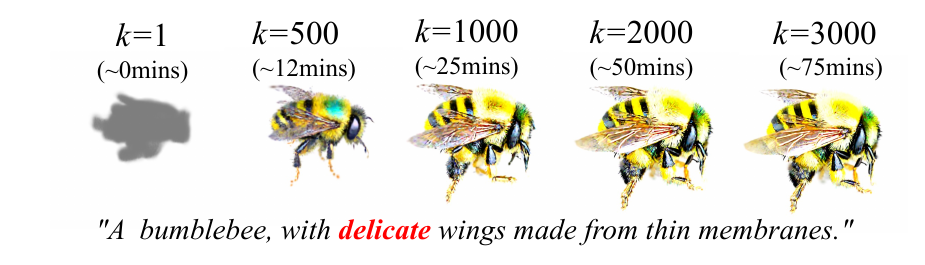}
\centering
\caption{The qualitative result of convergence speed in \sysname. $k$ represents the number of iterations.}
\label{fig:speed}
\end{figure}

\begin{figure}
\centering
\includegraphics[width=0.48\textwidth]{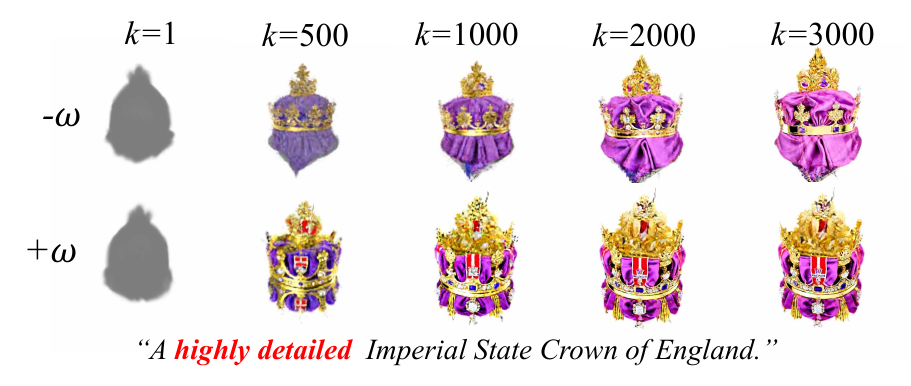}
\centering
\caption{The qualitative result of using DBC $\omega$ with \sysname.}
\label{fig:w}
\end{figure}

\noindent \textbf{Convergence Speed.}
\label{sec:time}
Figure~\ref{fig:convergence} compares the convergence speed of \sysname\ and baseline methods. It shows that our method achieves the fastest convergence while maintaining superior generation quality. In contrast, with continued optimization, methods such as GaussianDreamer~\cite{yi2023gaussiandreamer}, FPS~\cite{li2024dreamscene}, and LucidDreamer~\cite{liang2023luciddreamer} suffer from over-saturation, leading to a decline in generation quality, whereas \sysname\ steadily improves 3D generation results. Figure~\ref{fig:speed} presents qualitative results at different iteration steps $k$, demonstrating that \sysname\ can produce satisfactory results within $10\sim20$ minutes. This efficiency is attributed to the multi-granularity alignment, as discussed in Section~\ref{sec:GE3D}.

\par Figure~\ref{fig:convergence} and Figure~\ref{fig:w} further provide quantitative and qualitative evaluations of the impact of the dynamic balancing coefficient 
$\omega$ on generation time and quality. We find that 
$\omega$ further shortens convergence time while improving the overall generation quality by effectively regulating the transition of details, thus facilitating a more sequential and refined generation process.

\begin{figure}
\centering
\includegraphics[width=0.48\textwidth]{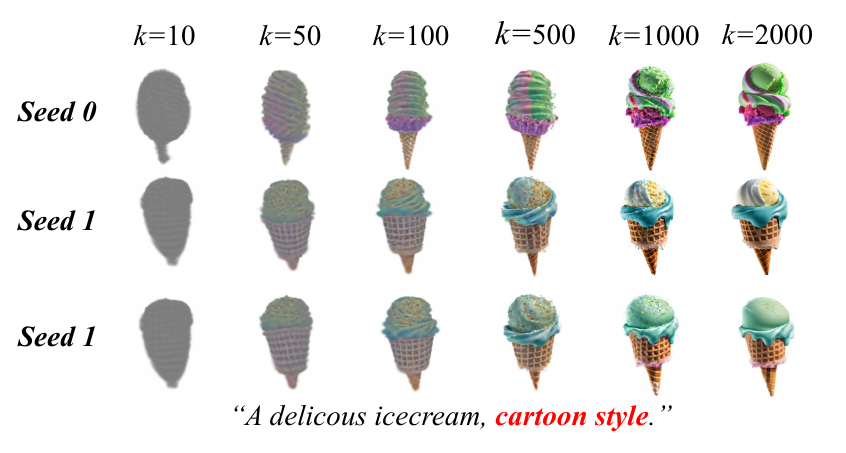}
\centering
\caption{Diversity results of \sysname.}
\label{fig:diversity}
\end{figure}
\subsection{Ablation Study and Analysis}
\noindent \textbf{Diversity.}
%
As shown in Figure~\ref{fig:diversity}, \sysname\ achieves excellent diversity. The same seed produces identical geometry during the coarse optimization stage but results in different textures in the fine optimization stage. In contrast, different seeds lead to variations in both geometry and texture. This is attributed to our structured multi-step generation process, as discussed in Section~\ref{sec:GE3D}.

\begin{figure}
\centering
\includegraphics[width=0.48\textwidth]{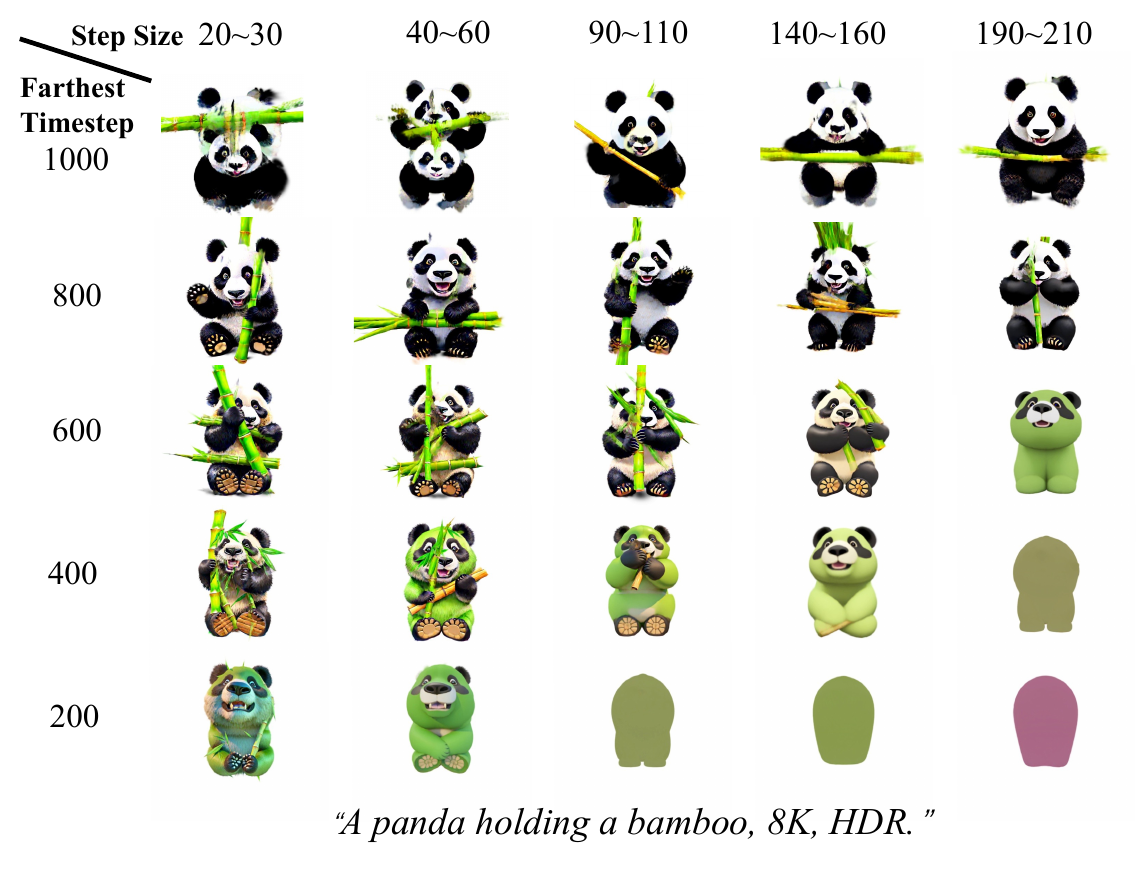}
\centering
\caption{Ablation result of the farthest timestep and  step sizes.}
\label{fig:ablation_t_n}
\end{figure}

\noindent\textbf{Impact of the Farthest Timestep and Step Size.} As shown in Figure~\ref{fig:ablation_t_n}, we conduct an ablation study on the impact of the farthest timestep and step size in the \sysname\ generation process, without using $\omega$ to adjust the proportion of different granularity information. The vertical axis represents the farthest timestep, while the horizontal axis indicates step size during each noising and denoising process. Our results highlight that the ratio of the farthest timestep to step size—denoted as the \textbf{\textit{number of steps 
$n$}}—is critical for 3D generation. When 
$n$ is too small, optimization fails (lower-right corner in Figure~\ref{fig:ablation_t_n}) due to insufficient guidance from the diffusion model. On the other hand, when 
$n$ is too large, the results degrade in quality (upper-left corner in Figure~\ref{fig:ablation_t_n}). Although more steps introduce finer details, excessive diffusion information destabilizes the optimization. From a 2D editing perspective, this occurs because the edited result diverges too much from the input image, reducing consistency. In 3D generation, this excessive divergence disrupts coherence and exacerbates the Janus (multi-head) problem. We find that around $5\sim10$ steps yield the best results (along the diagonal), striking a balance between stability and rich guidance. 

\noindent\textbf{Generate using only fine-grained information.} Notably, even using small step sizes with timesteps $0\sim200$ (lower-left corner in Figure~\ref{fig:ablation_t_n}), valid 3D representations emerge, challenging the belief in prior work (DreamTime~\cite{huang2023dreamtime}) that smaller step sizes cause instability. Thus, successful text-to-3D generation requires ensuring \textbf{\textit{sufficient}} guidance from the pre-trained diffusion model, achievable through \textbf{\textit{multi-step editing.}} However, relying solely on fine-grained information risks semantic inconsistency (e.g., a green panda). Therefore, selecting an optimal farthest timestep to balance coarse and fine-grained information is crucial for high-quality, semantically coherent 3D generation. \par Additional experiments and details are provided in Sections~\ref{sec:supp_middle} and~\ref{sec:supp_x_0} of the Supplementary Materials.


\section{Conclusion and Future Work}

In this paper, we analyzed the relationship between 3D generation and 2D editing, integrating them into a novel 3D framework. Unlike traditional SDS-based methods, which rely on a single-step noise difference, we supervised latents differences across the entire multi-step editing trajectory. This enables the distillation of rich semantic information from pretrained 2D diffusion models into 3D representations, yielding photorealistic results. Both theoretical analysis and experiments confirm the effectiveness of our method.
Our work bridges the gap between text-to-3D generation and 2D editing, paving the way for future research and the integration of advanced 2D editing techniques into 3D generation.

\noindent\textbf{Future Work.} We will further improve 3D generation quality by incorporating more advanced 2D editing methods, reducing the optimization time for fine-grained information, and enhancing generation efficiency.


{
    \small
    \bibliographystyle{ieeenat_fullname}
    \bibliography{main}
}

\clearpage
\setcounter{page}{1}
\maketitlesupplementary
\section{Preliminary}
\label{sec:supp_preliminay}
\par \textbf{Diffusion models}~\cite{ho2020denoising,song2020denoising,rombach2022high} generate data by estimating the denoising gradient direction of the current state distribution. During training, the model first adds $t$-timestep gaussian noise $\epsilon$ to the input data $x$ and get $x_t$:
\begin{equation}
    x_t = \sqrt{\bar{\alpha_t}}x + \sqrt{1-\bar{\alpha_t}}\epsilon . 
\label{eq_supp:ddpm}
\end{equation}
Then it learns a noise prediction network  $\phi$ by minimizing the loss between the predicted $\epsilon_{\phi}(x_t, t)$  noise and the actual noise $\epsilon$:
\begin{equation}
\mathcal{L}_t = \mathbb{E}_{x, \epsilon \sim \mathcal{N}(0, I)}\left[\left\lVert \epsilon_{\phi}(x_t, t) - \epsilon \right\rVert^2\right] .
\end{equation}
In the inference process, it predicts and removes noise step by step to complete the data generation.
\par \noindent \textbf{DDIM sampling}~\cite{song2020denoising} is used to estimate the relationship between any two noised latents, so it is often commonly employed to accelerate the inference process. We can transform $x_m$ to $x_k$ (no matter $m>k$ or $m<k$) using DDIM sampling (we set the coefficient of random perturbation $\delta_t = 0$ in ~\cite{song2020denoising}) by:
\begin{equation}
\setlength{\abovedisplayskip}{1pt}
\setlength{\belowdisplayskip}{1pt}
\resizebox{0.9\linewidth}{!}{
    $x_k = \sqrt{\overline{\alpha}_k} \frac{x_m - \sqrt{1-\overline{\alpha}_m}\epsilon_{\phi}(x_m,m)}{ \sqrt{\overline{\alpha}_m}} + \sqrt{1-\overline{\alpha}_k}\epsilon_{\phi}(x_m,m).$}
\label{eq_supp:ddim}
\end{equation}
\par \noindent \textbf{Classifier-Free Guidance (CFG)} ~\cite{ho2022classifier} is a technique used during generation to enable text-guided synthesis without relying on an explicit classifier. It introduces a mechanism where both unconditional and conditional generation collaboratively control the image generation. The formulation is as follows:
\begin{equation}
\resizebox{0.9\linewidth}{!}{
    $\tilde{\epsilon}_{\phi}(x_t,t,y,\emptyset) =  \epsilon_{\phi}(x_t,t,\phi) + \lambda(\epsilon_{\phi}(x_t,t,y) - \epsilon_{\phi}(x_t,t,\emptyset))$}, 
\label{eq_supp:CFG}
\end{equation}
where $y$ represents the conditional text embedding derived from the input prompt and $\emptyset$ represents the unconditional empty text embedding. $\lambda$ represents the guidance coefficient. A higher value of $\lambda$ increases conformity to the text 
$y$ but may decrease the quality of the generated images. 
\par \noindent\textbf{Score Distillation Sampling (SDS)}~\cite{poole2022dreamfusion} aims to distill information from a pretrained 2D text-to-image diffusion model. Initially, a camera pose $c$ is randomly selected to render a differentiable 3D representation~\cite{mildenhall2021nerf, shen2021deep, kerbl20233d} $\theta$ using rendering function $g(\theta,c)$ to obtain an image $x$. Random $t$-timestep noise is then added to $x$, and denoised by $\phi$. The 3D representation is updated by minimizing the difference between these two noises:
\begin{equation}
\resizebox{0.9\linewidth}{!}{
    $\nabla_{\theta}\mathcal{L}_{\text{SDS}}(\theta) = \mathbb{E}_{t,\epsilon,c}\left[f(t)(\tilde{\epsilon}_{\phi}(x_t, t, y, \emptyset) - \epsilon)\frac{\partial g(\theta,c)}{\partial \theta}\right]$},
\label{eq_supp:SDS}
\end{equation}
where $f(t)$ serves as a weighting function that adjusts based on the timesteps $t$. In SDS, the CFG coefficient is set to 100.
\section{Supplementary Methods}
\subsection{More connections between 2D editing and SDS text-to-3D methods}
\label{sec:supp_2dvs3d}
\subsubsection{Imagic vs. ProlificDreamer}
\textbf{Imagic}~\cite{kawar2023imagic} is a prominent  method in text-based 2D editing, with its editing process divided into three main stages. Initially, it aligns the target text embedding $y_{tgt}$ with a pretrained diffusion model~\cite{saharia2022photorealistic,rombach2022high} using a reconstruction loss  to ensure alignment with the input image. In the second stage, both the image and the optimized text embedding $y_{opt}$ are fixed, and the diffusion model undergoes fine-tuning through the reconstruction loss to enhance alignment. These two stages align the input image, text embedding and the diffusion model, facilitating the image inversion as discussed in Section~\ref{sec:related-editing}. The final stage focuses on editing: the interpolated target text embedding $y_{tgt}$ and optimized text embedding $y_{opt}$ are then fed into the fine-tuned diffusion model to produce the edited image. By aligning all three components in a high-dimensional space, Imagic ensures that the edited image captures both the content of the original image and the intent of the editing text, resulting in high-quality editing outcomes.
\par \noindent\textbf{ProlificDreamer}~\cite{wang2024prolificdreamer} utilizes a two-stage optimization in each iteration. In the first stage, it fixes the input image and  fine-tuned pretrained diffusion model by minimizing  $\mathbb{E}_{t,\epsilon,c}\left[ ||\epsilon_{\phi}(x_t,t,c,y)-\epsilon||_2^2\right]$. In the second stage, the parameters $\phi$ of the diffusion model  are frozen, and the 3D representation (input image) is optimized using the traditional SDS loss~\cite{poole2022dreamfusion}. 
\par \noindent \textbf{Comparing these two methods reveals that ProlificDreamer essentially conducts  an image editing process in each iteration.} Like in Imagic,  the first stage involves aligning the input image with the text embedding and the diffusion model. In the second stage, the input image is edited using the fine-tuned diffusion model and the target text embedding, optimizing towards the desired distribution for the text-guided rendered image. At this point, the noise vector from the pretrained model   $\epsilon_{pretrain}(x_t,t,y)$  balances the content of the input image with the editing text, ensuring a stable and high-quality optimization direction. Consequently, the resulting 3D model achieves excellent outcomes.

\subsubsection{Appropriate farthest timestep}

\par In \textbf{Prompt2Prompt}~\cite{hertz2022prompt}, it has been observed that introducing too much target text information during the editing process, without maintaining consistency with the original image, can cause the edited image to significantly deviate from the input. Conversely, prioritizing consistency too heavily may suppress the intended effects of the editing text, leading to inaccurate edits. This observation corresponds with our ablation experiments concerning the farthest timestep and step size. 
If the farthest timestep is too large or the number of steps is too high, consistency diminishes, destabilizing the optimization process and leading to the Janus (multi-head) problem, where the same text information is expressed in multiple  directions. On the other hand, excessively preserving consistency, such as using a very small farthest timestep and fewer steps, can yield results that do not align with the intended textual modifications. Therefore, we advocate for a \textbf{moderate} farthest timestep. This view is supported by findings from \textbf{DreamTime}~\cite{huang2023dreamtime}, which emphasize the importance of information provided at intermediate diffusion timesteps for effective 3D optimization.

\subsection{Easily combine with other methods.}
\label{sec:supp_adaptable}
\par \noindent \textbf{Perp-Neg}~\cite{armandpour2023re} addresses the Janus (multi-head) problem in SDS~\cite{poole2022dreamfusion} by using distinct text descriptions for different camera viewpoints (e.g., front, side, back views) as guidance. It employs negative text techniques to suppress undesired expressions from alternate viewpoints. The primary formula used in Perp-Neg is as follows: 
\begin{equation}
\left\{
\begin{aligned}
\epsilon_{\phi}^{pos_v} &= \epsilon_{\phi}(x_t,t,y_{pos}^{(v)}) -\epsilon_{\phi}(x_t,t,\emptyset)\\
\epsilon_{\phi}^{neg_v^{(i)}} &= \epsilon_{\phi}(x_t,t,y_{neg_{(i)}}^{(v)}) -\epsilon_{\phi}(x_t,t,\emptyset)\\
\tilde{\epsilon}_{\phi}^{PN}(x_t,t,v,y) &= \epsilon_{\phi}(x_t,t,\emptyset) + \omega[\epsilon_{\phi}^{pos_v}-\sum_{i}\omega_v^{(i)}\epsilon_{\phi}^{neg_v^{(i)\perp}}]
\end{aligned}
\right.
\end{equation}
where $y_{.}^{(v)}$ is the positive or negative text embedding at viewpoint $v$.  $\epsilon_{\phi}^{neg_v^{(i)\perp}}$ refers the perpendical component of $\epsilon_{\phi}^{neg_v^{(i)}}$ on the $\epsilon_{\phi}^{pos_v}$. And $\omega_v^{(i)}$ represent the $i$-th weights of negative prompts.
\par We can seamlessly integrate this method into our pipeline by substituting  $\tilde{\epsilon}_{\phi}(x_t,t,y,\emptyset)$  to $\tilde{\epsilon}_{\phi}^{PN}(x_t,t,v,y)$.
\par \noindent \textbf{Reconstructive generation} is a technique used in Formation Pattern Sampling (FPS)~\cite{li2024dreamscene} to  expedite the creation of surface textures. 
This method primarily employs a single-step denoising approach to generate images during the final stages of 3D representation optimization, where there is already good consistency. The process then effectively transforms the generation task into a reconstruction task. To enhance this, we replace the  \textbf{single-step} denoising with  \textbf{multi-step} denoising along the denoising trajectory, enabling the generation of edited images with detailed textures. We can similarly utilize reconstruction methods to further accelerate the generation process.

\begin{figure}
\centering
\subfloat[\label{fig:omega1}]{
		\includegraphics[width=0.24\textwidth]{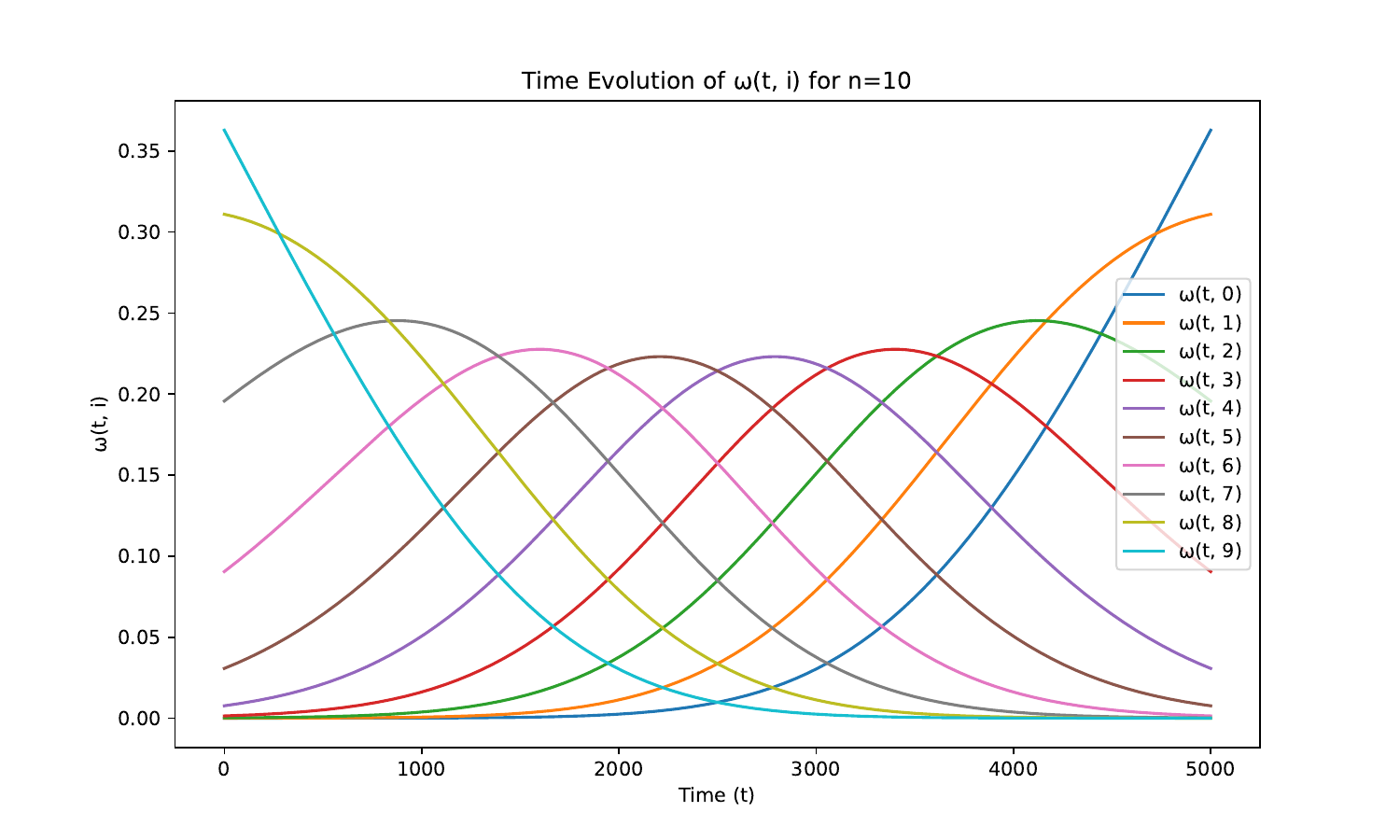}}
\subfloat[\label{fig:omega2}]{
\includegraphics[width=0.24\textwidth]{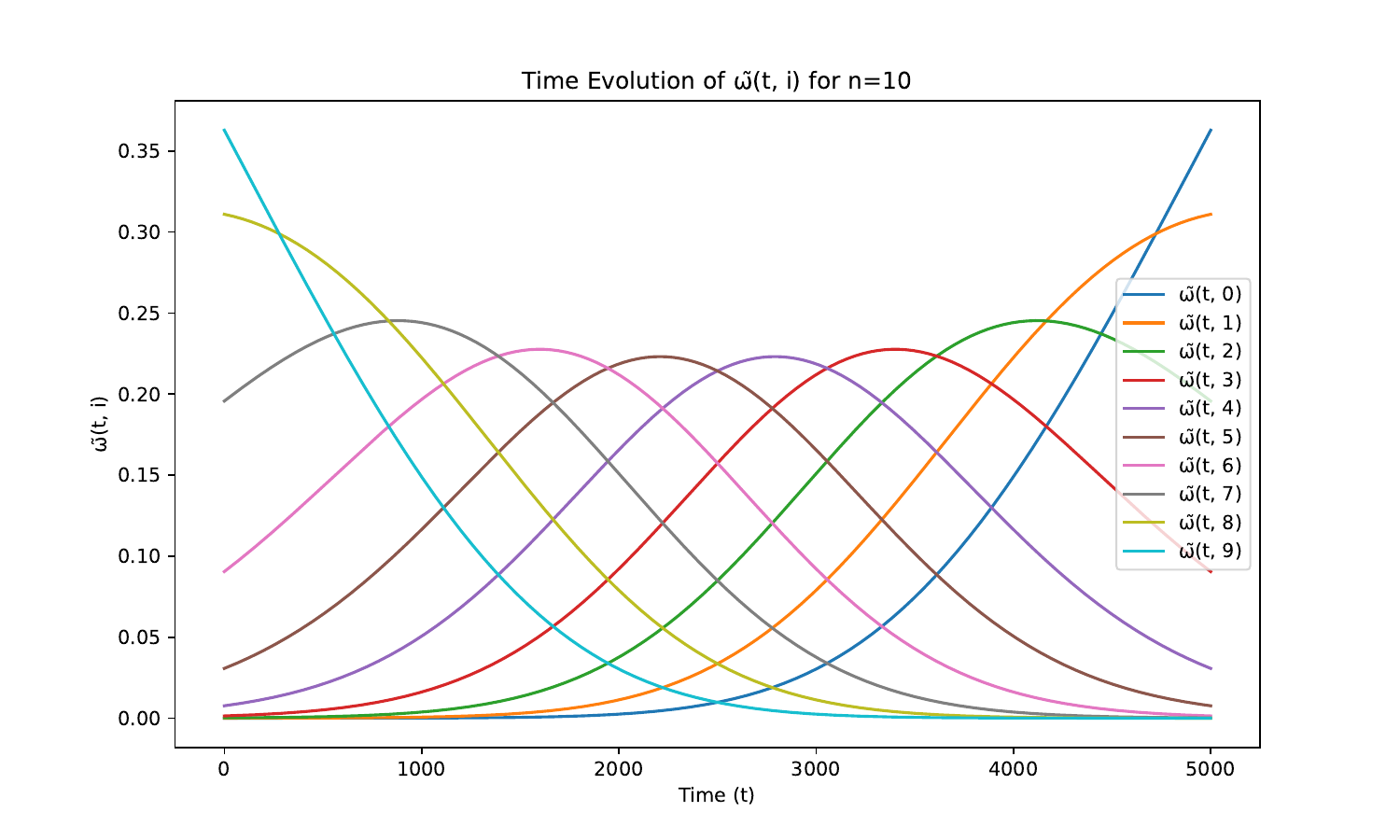}}
\vspace{-10pt}
\caption{(a) shows the $\omega(i,k)$ function plots for different granularities ($i$) and iteration counts ($k$), while (b) shows the normalized version of the plots.}
\label{fig:omega}
\vspace{-13pt}
\end{figure}

\section{Supplementary Experiments}
In this section, additional qualitative experimental results are provided in Section~\ref{sec:supp_more}. 
A more detailed qualitative comparison with ISM~\cite{liang2023luciddreamer} is available in Section~\ref{sec:supp_ISM}
The generation process of \sysname\ is showcased in Section~\ref{sec:supp_middle}, and an ablation study comparing the use of only $x_0 - \tilde{x}_0$ with \sysname\ is presented in Section~\ref{sec:supp_x_0}.

\subsection{Dynamic balancing coefficient}
\label{sec:supp_DBC}
We leverage the Gaussian function's characteristic of first increasing and then decreasing to facilitate the transition from coarse-grained to fine-grained information during optimization. As shown in Figure~\ref{fig:omega1}, we illustrate $10$ different Gaussian curves over $5000$ iterations, where the Gaussian peaks are evenly distributed across the total number of iterations. After normalization using Equation~\ref{eq:normalization}, as depicted in Figure~\ref{fig:omega2}, we achieve the transition of information at different granularities across various iteration steps.

\subsection{More results}
\label{sec:supp_more}
\begin{figure*}
\centering
\includegraphics[width=0.95\textwidth]{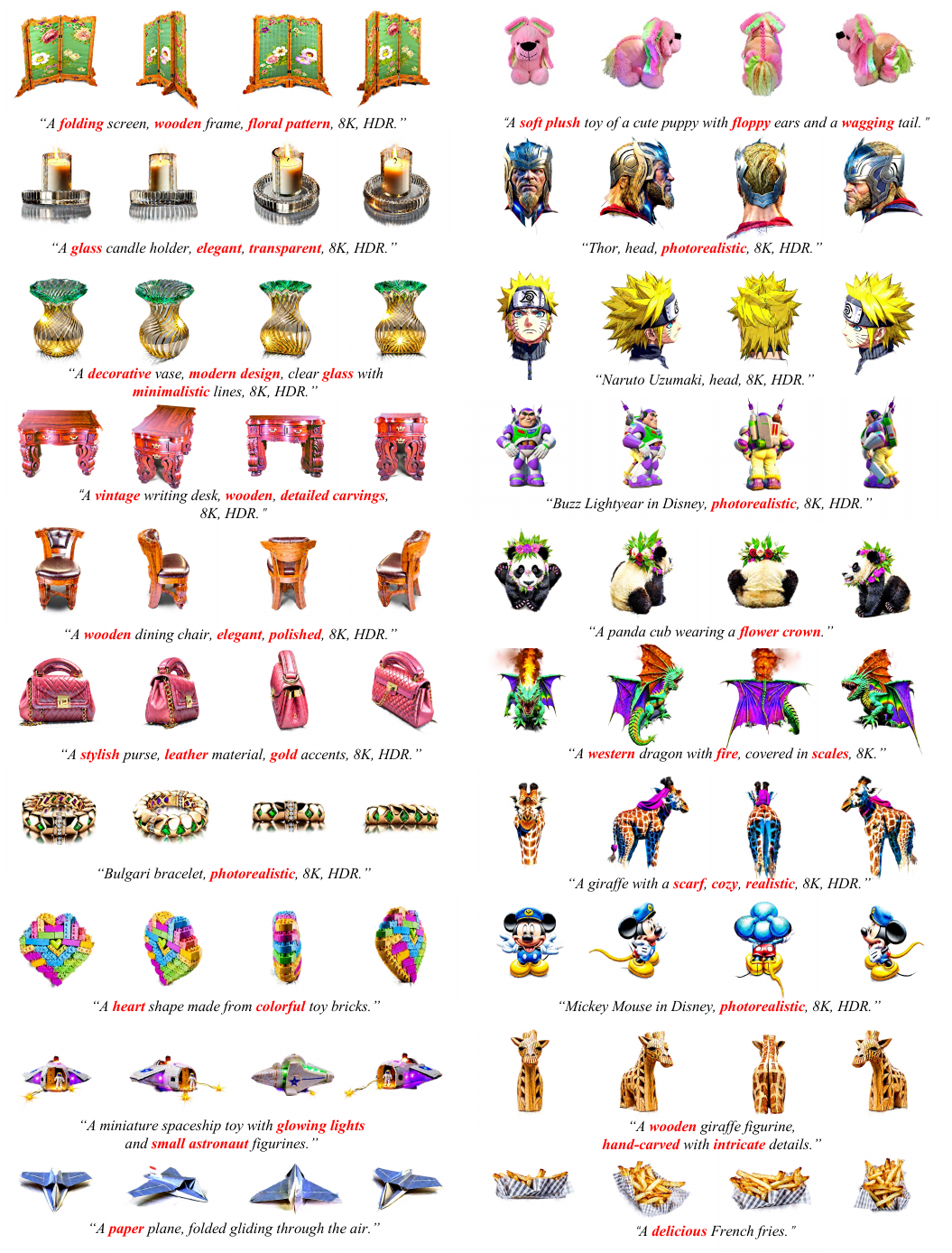}
\centering
\vspace{-10pt}
\caption{More quantitative results of \sysname.}

\label{fig:supp_teaser}
\end{figure*}

As demonstrated in Figures~\ref{fig:supp_teaser}, we provide  additional examples  of 3D generation from complex prompts in real-world scenarios. \sysname\ not only showcases its ability to produce photorealistic 3D content but also exhibits excellent generalization performance.
\subsection{Detailed Comparison with ISM}
\label{sec:supp_ISM}
\begin{figure*}
\centering
\subfloat[\textit{“Black Panther in Marvel, head, photorealistic, 8K, HDR.”}
]{
\includegraphics[width=\textwidth]{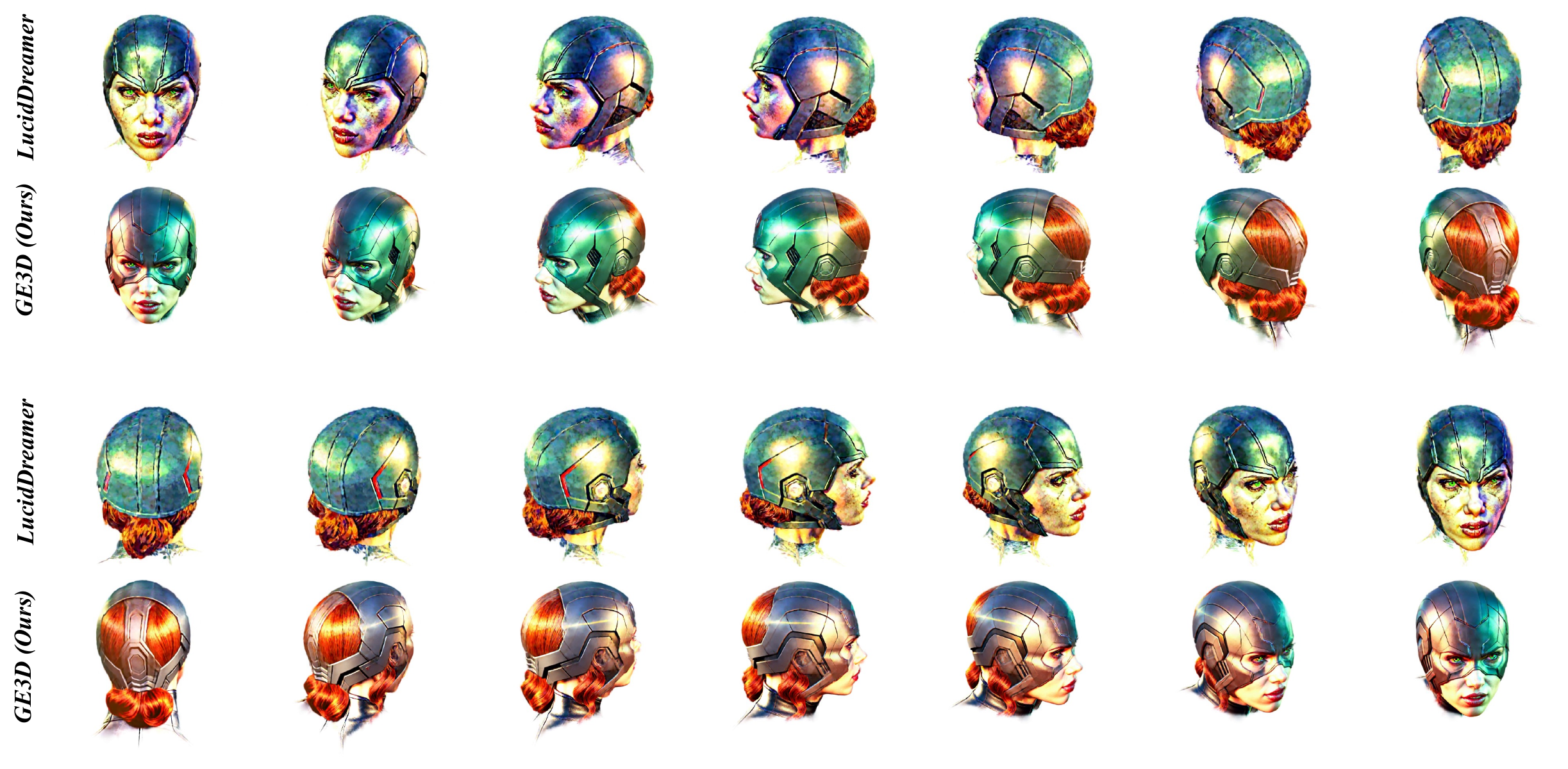}}
\\
\subfloat[\textit{“A delicious hamburger.”}]{
\includegraphics[width=\textwidth]{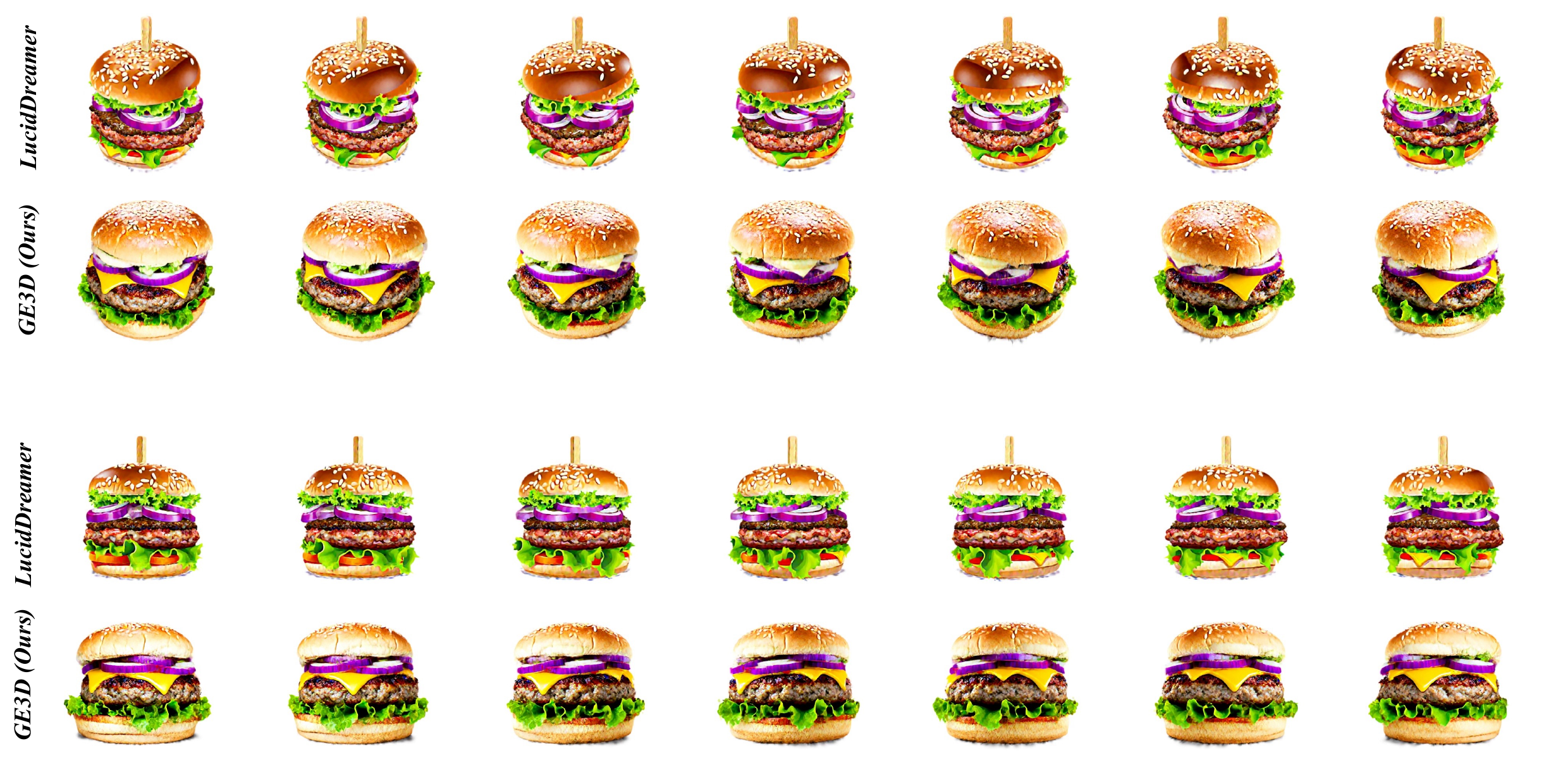}}

\caption{Detailed comparison between \sysname\ and ISM~\cite{liang2023luciddreamer}. The 3D content generated by \sysname\ is of higher quality and contains more detailed features.}
\label{fig:supp_lucid}
\end{figure*}

As shown in Figure~\ref{fig:supp_lucid}, we present a frame-by-frame comparison with ISM~\cite{liang2023luciddreamer} from multiple viewpoints. As quantitatively analyzed in Section~\ref{sec:why}, 
ISM utilizes a single-step editing process at a relatively large timestep to preserve coarse-grained information. 
Consequently, while its generated content aligns well with the text, it lacks fine-grained details. This deficiency is particularly noticeable in the textures of human faces and animal fur, where ISM struggles to produce ultra-high-quality 3D content.  By employing multi-step editing, our approach \sysname\ extracts information of varying granularities from the diffusion model, resulting in superior quality in the generated content.

\begin{figure}
\centering
\subfloat[\textit{"A panda holding a bamboo, realistic, 8K, HDR."}]{
\includegraphics[width=0.48\textwidth]{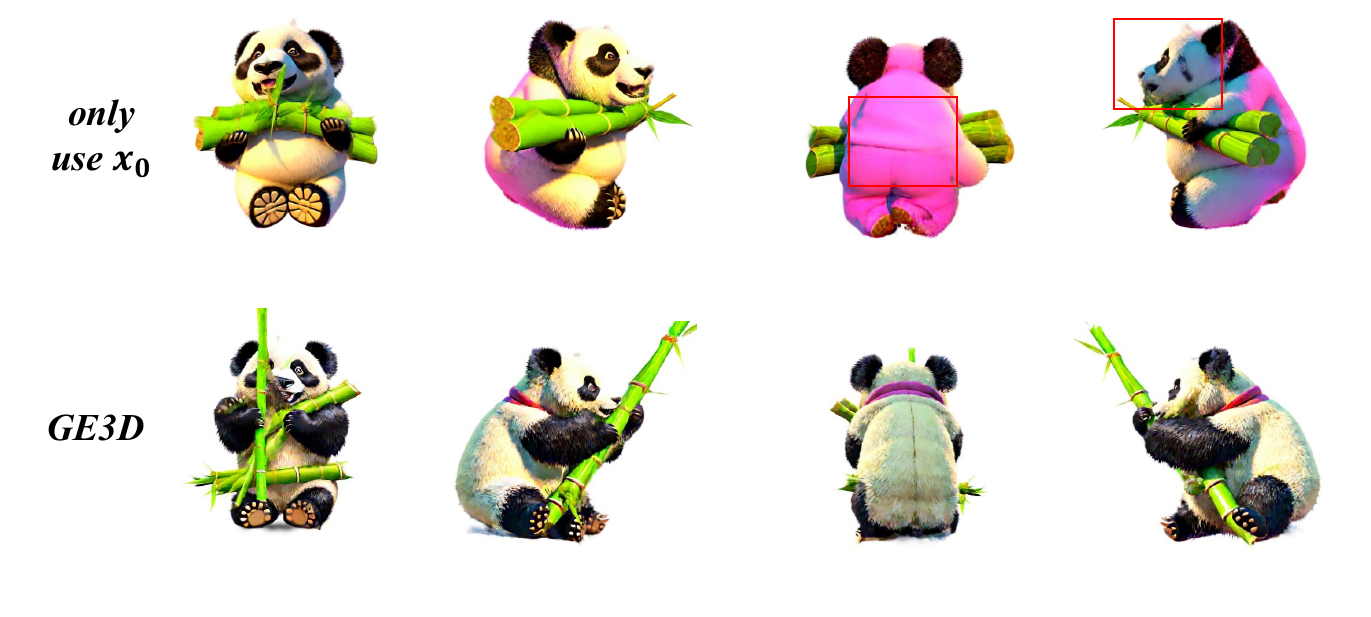}}
\\
\subfloat[\textit{"A DSLR of photo of a robot dinosaur."}]{
\includegraphics[width=0.48\textwidth]{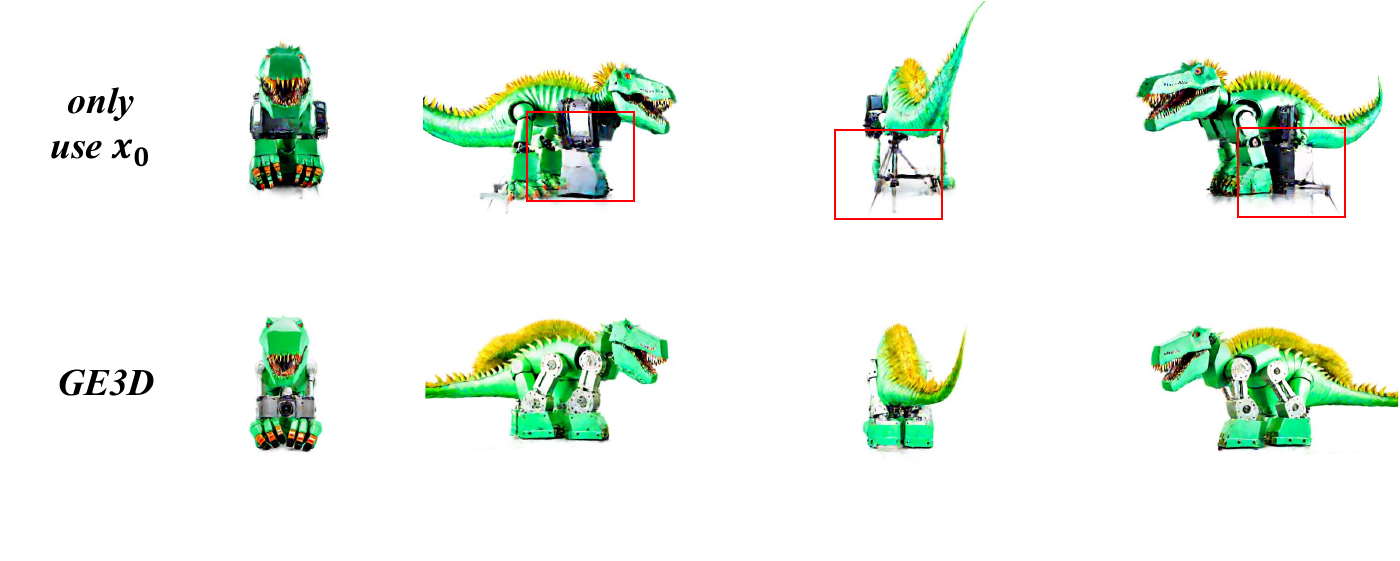}}

\caption{Comparison with methods using only $x_0 - \tilde{x}_0$. This method utilizes excessive fine-grained details, which can lead to the production of unreasonable content.}
\label{fig:supp_x_0}
\end{figure}

\begin{figure}
\centering
\includegraphics[width=0.48\textwidth]{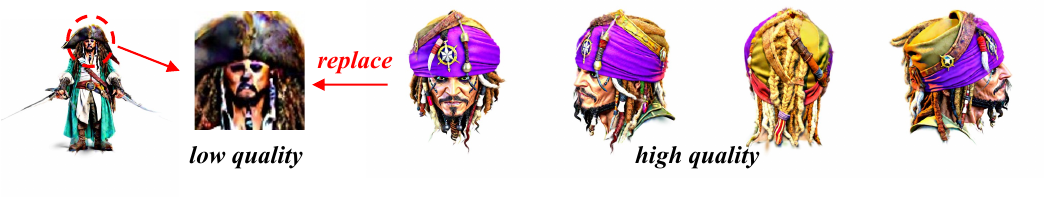}
\centering
\caption{The limitation of \sysname\ for generating the small head of avatar. It cannot generate as finely as when only generating a human head.}
\label{fig:limitation}
\end{figure}

\subsection{Generation Process of \sysname}
\label{sec:supp_middle}

\begin{figure*}
\centering
\includegraphics[width=\textwidth]{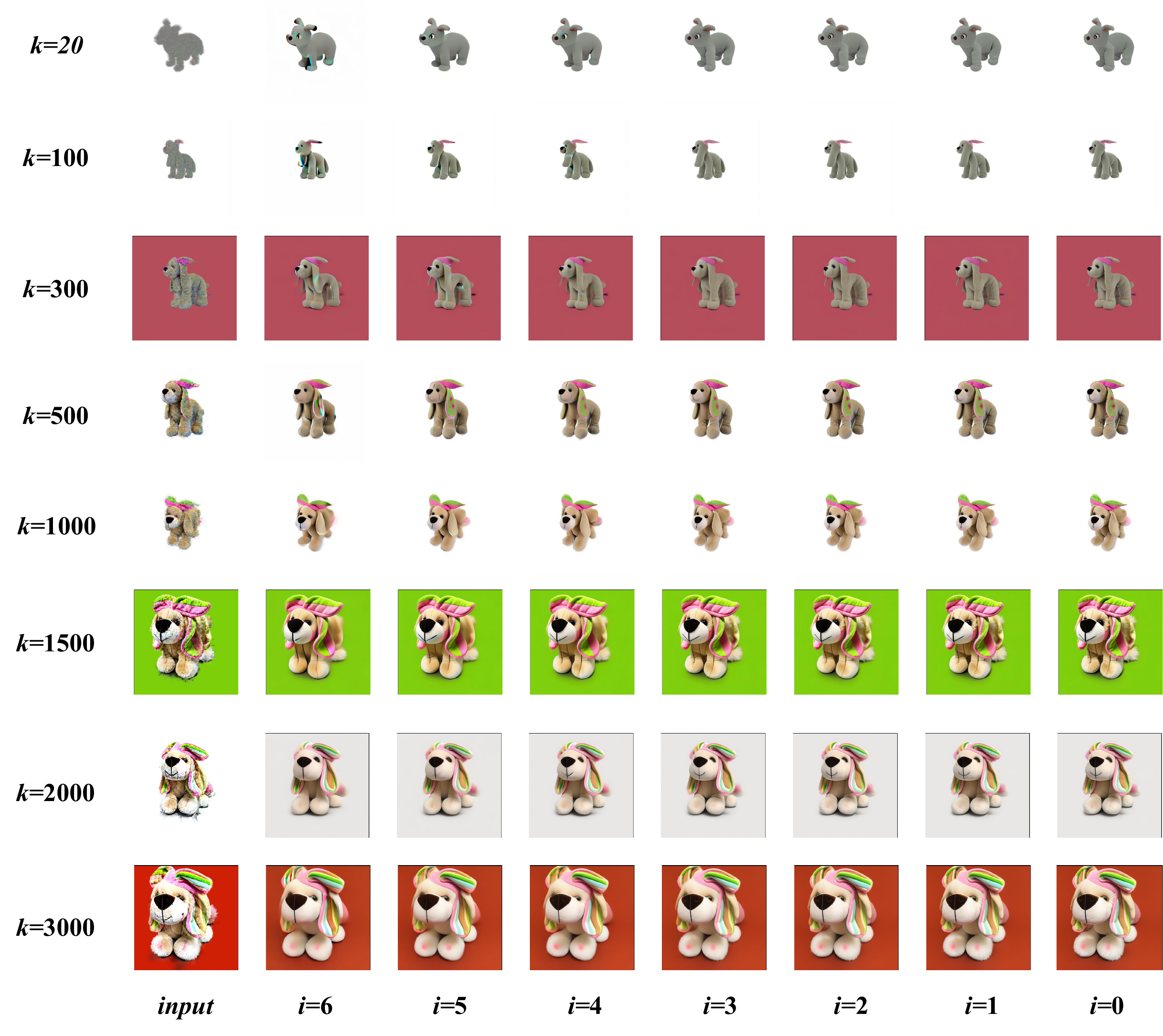}
\centering
\caption{The generation process of \sysname.}
\label{fig:supp_middle}
\end{figure*}

As illustrated  in Figure~\ref{fig:supp_middle}, we demonstrate the generation process of \sysname\ by displaying the results of pseudo-GT~\cite{liang2023luciddreamer} for images under different denoising diffusion trajectory steps $i$ and iterations $k$ (with a total of $3000$ iterations, $6$ diffusion steps, and a step size between $60$ and $80$). From left to right, the diffusion model initially emphasizes shape information at larger timesteps, resulting in simpler and smoother textures like fur. However, as shown on the far right, the model gradually provides more fine-grained details, such as realistic colors and textures. This progression supports our analysis in the main text that general content generation occurs within $500$ to $1000$ steps, with subsequent steps focused on refining surface textures to achieve higher-quality 3D results.
\subsection{Using only $x_0$}
\label{sec:supp_x_0}
As shown in Figure~\ref{fig:supp_x_0}, we evaluated the outcomes of optimizing the 3D representation using only $x_0 - \tilde{x}_0$. Despite requiring the same computational effort as our full multi-step editing process, this approach is more susceptible to the Janus problem—even with the incorporation of the Perp-Neg~\cite{armandpour2023re} module—and tends to produce unreasonable content. This issue arises because 
$\tilde{x}_0$
  retains excessive fine-grained details. At the initial generation stage, this can lead to the generation of the same textual content in multiple directions, resulting in irrational outcomes.
  
 \section{Limitations}
As shown in Figure~\ref{fig:convergence} of the main text, we spend more time generating fine-grained information. We plan to improve this process for better efficiency. Additionally, when generating avatars, the head often appears too small to be optimized with the same level of detail as the body, leading to blurry outputs as shown in Figure~\ref{fig:limitation}. To address this, we are developing a specialized camera sampling algorithm that ensures finer facial detail while optimizing the entire digital human.

\end{document}